\documentclass[runningheads]{llncs}

\usepackage{eccv}

\usepackage{eccvabbrv}

\usepackage{graphicx}
\usepackage{booktabs}

\usepackage[accsupp]{axessibility}  

\usepackage{algorithm}
\usepackage{algorithmicx}
\usepackage{algpseudocode}
\usepackage{multirow} 
\usepackage{amssymb,bm}
\usepackage{soul}
\usepackage{color}
\usepackage{tabularx}
\usepackage{marvosym}

\usepackage{hyperref}

\usepackage{orcidlink}

\makeatletter
\def\thanks#1{\protected@xdef\@thanks{\@thanks
        \protect\footnotetext{#1}}}
\makeatother

\hypersetup{
    colorlinks=true
}

\begin{document}
	
\def \x {\bm{x}}
\def \y {\bm{y}}
\def \p {\bm{p}}
\def \z {\bm{z}}
\def \q {\bm{q}}
\def \l {\bm{l}}
\def \g {\bm{g}}

\title{Dual-Decoupling Learning and Metric-Adaptive Thresholding for Semi-Supervised Multi-Label Learning} 

\titlerunning{D2L and MAT for SSMLL}

\author{ Jia-Hao Xiao * \inst{1}\orcidlink{0009-0005-4363-290X} \and
Ming-Kun Xie *\thanks{* Equal Contribution} \inst{1}\orcidlink{0000-0002-1053-1409} \and
Heng-Bo Fan \inst{1}\orcidlink{0009-0000-3910-8180} \and
Gang Niu \inst{2}\orcidlink{0000-0002-7353-5079} \and
\\ Masashi Sugiyama \inst{2,3}\orcidlink{0000-0001-6658-6743} \and
Sheng-Jun Huang \textsuperscript{\Letter}\thanks{\textsuperscript{\Letter} Corresponding Author} \inst{1}\orcidlink{0000-0002-7673-5367}}

\authorrunning{J.H. Xiao, M.K. Xie et al.}

\institute{Nanjing University of Aeronautics and Astronautics, Nanjing, China \and
RIKEN Center for Advanced Intelligence Project, Tokyo, Japan \and
The University of Tokyo, Tokyo, Japan\\
\email{\{jiahaoxiao,mkxie,fan\_heng\_bo,huangsj\}@nuaa.edu.cn}\\
\email{gang.niu.ml@gmail.com} \ \ \email{sugi@k.u-tokyo.ac.jp}}
\maketitle

\begin{abstract}
  Semi-supervised multi-label learning (SSMLL) is a powerful framework for leveraging unlabeled data to reduce the expensive cost of collecting precise multi-label annotations. Unlike semi-supervised learning, one cannot select the most probable label as the pseudo-label in SSMLL due to multiple semantics contained in an instance. To solve this problem, the mainstream method developed an effective thresholding strategy to generate accurate pseudo-labels. Unfortunately, the method neglected the quality of model predictions and its potential impact on pseudo-labeling performance. In this paper, we propose a dual-perspective method to generate high-quality pseudo-labels. To improve the quality of model predictions, we perform dual-decoupling to boost the learning of correlative and discriminative features, while refining the generation and utilization of pseudo-labels. To obtain proper class-wise thresholds, we propose the metric-adaptive thresholding strategy to estimate the thresholds, which maximize the pseudo-label performance for a given metric on labeled data. Experiments on multiple benchmark datasets show the proposed method can achieve the state-of-the-art performance and outperform the comparative methods with a significant margin. The implementation is available at \href{https://github.com/JiahaoXxX/SSMLL-D2L_MAT}{JiahaoXxX/SSMLL-D2L\_MAT}.
\end{abstract}

\section{Introduction}
\label{sec:introduction}

Multi-label learning (MLL) is a mainstream and practical framework for dealing with the problem where each object is associated with multiple class labels \cite{zhou2012multi}. For example, a seascape image may be annotated with labels \textit{sea}, \textit{sunset} and \textit{beach}. MLL has achieved great successes in many research areas, \eg, music emotion recognition \cite{zhang2021multi}, text categorization \cite{lin2018semantic} and image annotation \cite{chen2019multi}. 

However, in real-world scenarios, obtaining precise annotations for every instance can be time-consuming and expensive. This challenge becomes significantly more daunting, especially when dealing with a vast output space, leading to exponential increases in complexity. Semi-supervised multi-label learning (SSMLL) has gained increased attention in recent years as it demands fewer annotations for training, making it increasingly appealing for various applications \cite{xie2023class,wang2020dual,tan2017semi,li2021online}. In SSMLL, only a limited portion of the training data is fully annotated, while the majority of the training data remains entirely unlabeled. The goal is to boost the model performance by exploiting the information of unlabeled examples.

To harness the potential of unlabeled examples, the most prevalent method is pseudo-labeling, which generates the pseudo-labels for unlabeled examples based on the model trained with labeled ones. The effectiveness of pseudo-labeling has been demonstrated by many excellent methods in both single-label learning scenario, \ie, semi-supervised learning (SSL) \cite{sohn2020fixmatch,wang2023freematch,zhang2021flexmatch,guo2022class}, and MLL scenario \cite{xie2023class,wang2020dual,li2021online}. Unlike SSL that often selects the largest probable label as the pseudo-label, the variability of unknown label count for each unlabeled instance in SSMLL presents a significant challenge on obtaining accurate pseudo-labels. To address this problem, the recently proposed method class-aware pseudo-labeling \cite{xie2023class} performs pseudo-labeling in a class-wise manner, which generates pseudo-labels for unlabeled instances on a class-by-class basis. This method transforms a hard problem of estimating the label count of each unlabeled instance into a far easier problem of estimating the class proportions of unlabeled data. The transformation ultimately improves the pseudo-labeling performance. Although the method has made progress in SSMLL, it merely focuses on capturing the true class proportions, while neglecting the quality of model predictions, which is also an essential factor of achieving favorable pseudo-labeling performance.

In this paper, we propose a novel approach to solve the SSMLL problem, which aims to generate high-quality pseudo-labels from dual perspectives. To improve the quality of model predictions, we propose the Dual Decoupling Learning (D2L) method to respectively decouple correlative/discriminative features and the generation/utilization of pseudo-labels. The former exploits co-occurrence relationships among different labels in a whole image and captures distinctive patterns for each individual object in different patches; the later uses two classification heads to generate and utilize pseudo-labels independently, which prevents the accumulation of incorrect pseudo-labels. To achieve proper separations between positive and negative labels, we develop the Metric-Adaptive Thresholding (MAT) strategy, to estimate the thresholds, which achieves the optimal pseudo-label performance in terms of an evaluation metric based on the labeled data. Extensive experiments on benchmark datasets validate that the proposed method can achieve the state-of-the-art performance.


To sum up, the contributions of this work are as follows:
\begin{itemize}
	
	
	\item The proposed D2L approach performs dual decoupling to enhance the learning of correlative and discriminative features, while simultaneously refines the generation and utilization of pseudo-labels.
	
	
	\item The proposed MAT strategy estimates the threshold per class in a metric-adaptive manner, with the goal of achieving the optimal pseudo-labeling performance in terms of a given measurement. 
	
	
	\item The proposed method obtains the state-of-the-art results with a significant margin across multiple datasets and labeled proportions. For example, on the benchmark multi-label dataset COCO \cite{LinMBHPRDZ14}, our method improves upon the state-of-the-art performance by 4.2\% -- 6.9\% mAP score on different labeled proportions. 
	
\end{itemize}

\section{Related Works}
\label{sec:related}
In this section, we introduce related works, including MLL, multi-label learning with missing labels (MLLML), and SSMLL.
\subsection{Multi-Label Learning}
In MLL, each instance is associated with multiple class labels. The mainstream research direction in this field is modeling correlations between labels \cite{xie2024counterfactualreasoningmultilabelimage,huang2012multi,zhu2017multi,zhang2018binary}. Graph convolution networks (GCN) are usually utilized to capture the co-occurrence among labels, \eg, ML-GCN \cite{chen2019multi} and ADD-GCN \cite{ye2020attention}. Besides, recurrent models \cite{wang2016cnn,yazici2020orderless} and transformer encoder structures \cite{dosovitskiy2020image,touvron2021training,liu2021swin} have also been used to construct complementary relationships in MLL. Another research direction is to deal with the imbalance issue, including inter-class (long-tail) \cite{guo2021long} and intra-class (positive-negative) \cite{ridnik2021asymmetric} imbalance. \cite{guo2021long} developed a two-branch network for collaborative learning on both uniform and re-balanced samplings. \cite{ridnik2021asymmetric} proposed that asymmetric loss (ASL) which is a representative method that dynamically down-weights and hard-thresholds easy negative examples. Moreover, considering that an image often contains multiple objects, the methods designed to locate areas of interest related to semantic labels by using attention technique arose \cite{zhu2017learning,chen2019learning}. All the aforementioned methods are based on the fully supervised setting. However, acquiring fully labeled data is far more challenging and time-consuming compared to obtaining partially labeled or unlabeled data in the current context of rapidly growing big data. Thus, some weakly-supervised multi-label learning settings have emerged recent years, \ie, to train the model with incomplete, inexact or inaccurate data, like semi-supervised multi-label learning \cite{wang2020dual,xie2023class}, multi-label learning with missing labels \cite{durand2019learning,ben2022multi,huynh2020interactive}, partial multi-label learning \cite{xie2018partial,zhang2020partial}, multi-label learning with noisy labels \cite{li2022estimating,hu2019weakly} and so on.

\subsection{Multi-Label Learning with Missing Labels}
In practice, the high cost of collecting precise multi-label annotations is always a trouble, which turns the research attention to train with limited labeled information. Multi-label learning with missing labels (MLLML) \cite{durand2019learning,ben2022multi,huynh2020interactive} is a research direction that effectively reduces annotation costs, where each example is partially annotated. There are many large-scale datasets resorting to partial annotations \cite{liu2021emerging}, \eg, OpenImages \cite{kuznetsova2020open} and LVIS \cite{gupta2019lvis}, where only a small fraction of labels are annotated. Moreover, many excellent methods have already been proposed to address this issue. \cite{durand2019learning} proposed the partial Binary Cross-Entropy (BCE) that normalizes the loss by the proportion of known labels. \cite{ben2022multi} individually addressed each unannotated label by considering two probabilistic measures: label likelihood and label prior. 
Even in an extreme MLLML scenario where each example has only one annotated positive label, there have also been many outstanding works emerging in recent years. \cite{ColeALPMJ21} treated all unobserved labels as negatives and employed regularization to mitigate the adverse effects of false negative labels. \cite{kim2022large} empirically observed the memorization effect, then proposed to reject or correct the large loss examples.






\subsection{Semi-Supervised Multi-Label Learning}
Despite the existence of numerous approaches to address SSMLL, most of them are based on training linear models \cite{chen2008semi,guo2012semi,wang2013dynamic,zhao2015semi,zhan2017inductive,tan2017semi}. With deep learning models demonstrating formidable learning capabilities, the integration of deep learning has become a trend. However, SSMLL has been relatively underexplored in the field of deep learning in recent years. \cite{wang2020dual} firstly introduced the deep model into SSMLL, and proposed a novel approach which jointly explores the feature distribution and the label relation simultaneously. \cite{xie2023class} proposed class-aware thresholds to effectively control the assignment of positive and negative pseudo-labels for each class. Thanks to the similarity in problem setups, methods for handling unlabeled data in SSL can, to some extent, be adapted to SSMLL. Pseudo-labeling has been extensively developed and employed for harnessing unlabeled data, as a mainstream approach in SSL. \cite{guo2022class} proposed an adaptive thresholding strategy for different classes under class-imbalanced distributions, which only required a fixed threshold for the most majority category, while the thresholds for other categories were determined based on the selection proportion of the most majority category. \cite{wang2023freematch} adopted both global and local self-adaptive thresholds computed from the exponential moving average (EMA) of prediction statistics from unlabeled examples. The previous methods always tried to design effective  pseudo-label selection strategies, but overlooked the potential influence of model outputs on pseudo-labels. Both factors should jointly determine the quality of pseudo-labels.





\section{The Method}
\label{sec:method}
In this section, we first present some preliminary concepts, such as symbol representation and loss functions; subsequently, we will elaborate on our proposed dual-decoupling learning (D2L) framework and metric-adaptive thresholding (MAT) strategy.

\subsection{Preliminaries}
Let $\x\in\mathcal{X}$ be a feature vector and $\y\in\mathcal{Y}$ be a label vector, where $\mathcal{X}\in\mathbb{R}^d$ is an input space and $\mathcal{Y} = \{0,1\}^K$ is a target space with $K$ possible class labels. Here, $y_k=1$ indicates the $k$-th label is relevant to $\x$; $y_k=0$, otherwise. 
In SSMLL, we are given a labeled dataset $\mathcal{D}_L = \{(\x_i, \y_i)\}^N_{i=1}$ and an unlabeled dataset $\mathcal{D}_U = \{\x_j\}^M_{j=1}$, where $N$ and $M$ are the numbers of labeled and unlabeled training examples, which satisfy $M\gg N$. Let $\p_i$ and $\q_i$ be the predicted probabilities on strongly-augmented instance $\text{Aug}_\mathrm{s}(\x_i)$ and weakly-augmented instance $\text{Aug}_\mathrm{w}(\x_i)$, where $\text{Aug}_\mathrm{w}(\cdot)$ and $\text{Aug}_\mathrm{s}(\cdot)$ represent stochastic weak and strong augmentations. We will discuss how to obtain these predicted probabilities latter.


Regarding the labeled training examples, the most commonly used strategy is to aggregate a binary loss between model predictions and true labels over $K$ classes
\begin{equation}
	\label{eq:lb_loss}
	\mathcal{L}_L=\sum_{i=1}^N\sum_{k=1}^K\ell(p_{ik},y_{ik}).
\end{equation}
Here, we use the ASL \cite{ridnik2021asymmetric} for $\ell(\cdot, \cdot)$, since it often achieves better performance than the commonly used BCE loss. 

Regarding the unlabeled training examples, we aggregate the ASL between model predictions and pseudo-labels over $K$ classes
\begin{equation}
	\label{eq:unlabeled}
	\mathcal{L}_U=\sum_{j=1}^M\sum_{k=1}^K\ell(p_{jk},\hat{y}_{jk}),
\end{equation}
where $\hat{y}_{jk}=\mathbb{I}(q_{jk}\geq\tau_k)$ represents the pseudo-label of an unlabeled instance $\x_j$ and $\tau_k$ is the threshold for the class $k$. The overall objective function is defined as
\begin{equation}\label{eq:training}
	\mathcal{L}=\mathcal{L}_L+\mathcal{L}_U.
\end{equation}
Obviously, high-quality pseudo-labels often lead to favorable model performance. As mentioned before, the pseudo-labeling performance depends on two elements,  the thresholds $\tau_k$ and model predictions $\q_j$. Below, we will improve the quality of pseudo-labels from two perspectives: 1) develop the dual-decoupling learning framework to obtain accurate model predictions; 2) design the metric-adaptive thresholding method to acquire proper thresholds.

\subsection{Dual-Decoupling Learning}
\label{sec:D2L}
\paragraph{Correlative and Discriminative Features Decoupling} Many existing works \cite{chen2019multi,lanchantin2021general} focus on exploiting the label correlations, which have been regarded as an essential information for MLL. Unfortunately, excessively focusing on label correlations, \eg, co-occurrence, often leads to overlooking discriminative information for individual objects, making it hard for the model to recognize indistinguishable targets, \eg, tiny objects. Therefore, there exists a trade-off between considering co-occurrence relationships and capturing discriminative information. To address this problem, we propose to decouple feature learning into two sub-tasks, correlative feature learning, which exploits co-occurrence patterns among different objects, and discriminative feature learning, which captures the distinctive pattern of each individual object. These two types of information complement each other, collectively enhancing the model's recognition capabilities and consequently improving pseudo-labeling performance significantly. 

Specifically, for each training instance, we first crop it into $n$ patches, \ie, $\left\{\z^{(i)}_o\right\}_{o=1}^n=\text{Crop}(\x_i)$ of equal size, where $\text{Crop}(\cdot)$ is the cropping operation; by applying strong augmentation to the instance and patches and feeding them into the backbone $f(\cdot)$, \eg, ResNet \cite{he2016deep}, we obtain the correlative features $\g_i=f(\x_i)$ and discriminative features $\l^{(i)}_o=f(\z_o^{(i)})$ for each patch. Importantly, $\g_i$ encodes the correlative information of a whole image, while $\l^{(i)}_o$ captures the discriminative information of each patch. Then, we use a dual-head classifier consisting of a correlative classification head $h^g(\cdot)$ and a discriminative classification head $h^l(\cdot)$ to obtain the predicted probabilities $\p_i^g=h^g(\g_i)$ and $\p^{(i)}_o=h^l(\l^{(i)}_o)$. 

To merge the information from different patches, we aggregate the probabilities $\p^{(i)}_o$ into the local probability $\p_i^l$ in a spatially-weighted manner as 
\begin{equation}
	\label{eq:local_sum}
	\p_i^l=\sum_{o=1}^n\frac{\exp(\p^{(i)}_o/\alpha)}{\sum_{o=1}^n\exp(\p_o^{(i)}/\alpha)}\cdot\p^{(i)}_o,
\end{equation}
where $\alpha$ is a temperature parameter that controls the extent of focusing on a specific location. For any instance $\x_i$, we obtain the final prediction by averaging the correlative and discriminative probabilities $\p_i=(\p^g_i+\p^l_i)/2$. Moreover, we can obtain the predicted probabilities $\q_i^g$, $\q_i^l$, and $\q_i$ by applying weak augmentation to the instance and patches.

\begin{figure}[t]
	\centering
	\includegraphics[width=0.99\linewidth]{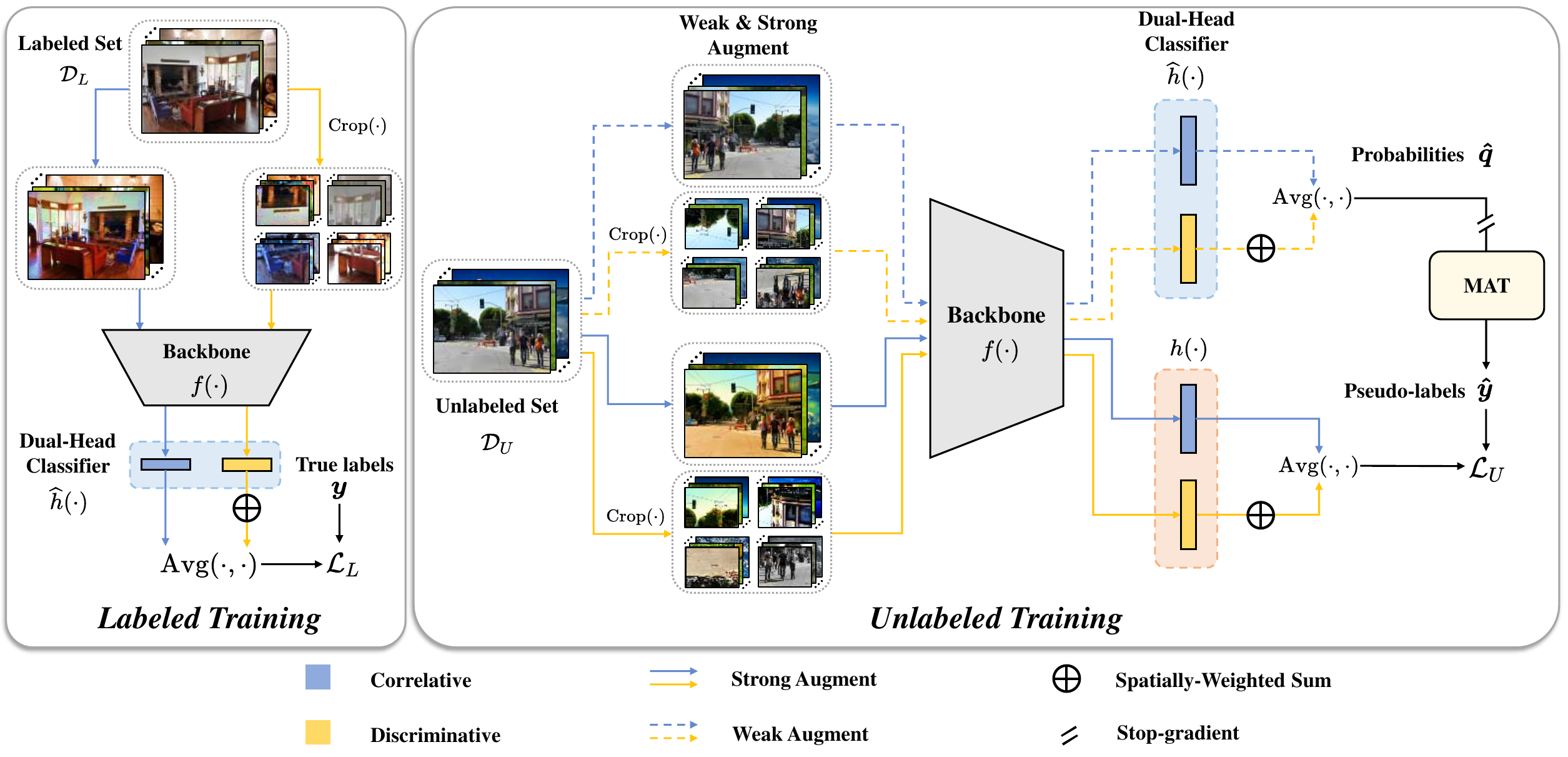}
	
	\caption{An illustration of the proposed learning framework. Blue and yellow colors represent correlative-wise and discriminative-wise components, while solid and dashed lines denote the strong and weak data augmentation streams, respectively. `Spatially-Weighted Sum' indicates the aggregation of probabilities from patches (see \Cref{eq:local_sum}). The pseudo-label generation process is gradient-free.}
	\label{fig:framework}
\end{figure}

\paragraph{Generation and Utilization of Pseudo-Labels Decoupling} One potential limitation of the previous work \cite{xie2023class} is that it trains the model using pseudo-labels generated by the same model. Although the method employs the teacher-student training strategy and consistency regularization to alleviate over-fitting, it often leads the model to accumulate error once incorrect pseudo-labels are generated. To address this issue, inspired by the previous work \cite{chen2022debiased}, we propose to decouple the generation and utilization of pseudo-labels, which aims to avoid the risk of accumulating pseudo-labeling errors. Similar to \cite{lee2021abc,wang2023imbalanced}, which introduce auxiliary classifiers to divert different data streams and alleviate confirmation bias, we use two dual-head classifiers $\{\widehat{h}^g(\cdot),\widehat{h}^l(\cdot)\}$ and $\{h^g(\cdot),h^l(\cdot)\}$ to generate and utilize pseudo-labels independently. On one hand, we train the classifiers $\{\widehat{h}^g(\cdot),\widehat{h}^l(\cdot)\}$ in an unbiased manner by inducing them on the clean labeled examples. Since the classifiers are trained on fully clean examples, we expect them to generate high-quality pseudo-labels. On the other hand, we train the classifiers $\{h^g(\cdot),h^l(\cdot)\}$ on pseudo-labels, which aim to boost feature learning by making full use of unlabeled examples and alleviate over-fitting to labeled ones. 

%
To generate pseudo-labels, for each unlabeled instance $\x_j$, we define $\hat{\q}_j=(\hat{\q}_j^g+\hat{\q}_j^l)/2$ as the probabilities predicted by $\{\widehat{h}^g(\cdot),\widehat{h}^l(\cdot)\}$. Then, we obtain the pseudo-label as 
\begin{equation}
	\label{eq:pseudo_label}
	\forall k\in[K],\quad\hat{y}_{jk}=\mathbb{I}(\hat{q}_{jk}\geq\tau_k),
\end{equation}
where $\tau_k$ is the threshold for class $k$. Then, we train the classifiers $\{h^g(\cdot),h^l(\cdot)\}$ based on the unlabeled data with \Cref{eq:unlabeled}. \Cref{fig:framework} provides an illustration of the dual-decoupling learning framework. In \Cref{sec:MAT}, we will introduce how to obtain proper thresholds.

\subsection{Metric-Adaptive Thresholding}
\label{sec:MAT}
As mentioned before, in SSMLL, we cannot directly select the most probable label as the pseudo-label for each unlabeled instance as commonly done in SSL. The recent method CAP \cite{xie2023class} solved this problem by generating pseudo-labels in a class-wise manner and using the empirical class proportions of labeled examples as approximate estimates of the true ones of unlabeled examples. Although CAP can precisely capture the true class proportions, it may not achieve optimal pseudo-labeling performance due to imperfect predictions. If the predicted probabilities are not entirely precise, meaning that a part of negative labels have greater probabilities than positive ones, it will introduce incorrect pseudo-labels. In other words, higher-ranked negative labels will be mislabeled as positive, while lower-ranked positive labels will be mislabeled as negative. The phenomenon will occur especially when the number of labeled examples is small and the model performance is limited.


To address this problem, instead of estimating the class proportions of unlabeled data, we propose the metric-adaptive thresholding (MAT) strategy to directly estimate the optimal threshold for every class. 
Specifically, we first define $\mathcal{M}(\hat{Y}_k, Y_k)$ as the metric function for evaluating class-wise pseudo-labeling performance, where $\hat{Y}_k$ and $Y_k$ represent the pseudo-labels and true labels for the $k$-th class. The intuition behind MAT is that given the metric $\mathcal{M}$, the optimal threshold is the one that maximizes the metric for every class. Unfortunately, without the access to true labels, it is difficult to directly obtain the optimal thresholds on the unlabeled data. Alternatively, a feasible solution is to estimate the thresholds on labeled data. Then, we can obtain thresholds based on the following thresholding strategy:
\begin{equation}
	\label{eq:taok_star}
	\forall k\in[K],\quad\tau^{\star}_k={\arg\max}_{\tau_k\in[0,1]}\mathcal{M}(\hat{Y}_k,Y_k),
\end{equation}
where $\hat{Y}_k=\{\hat{y}_{ik}\}_{i=1}^N$ and $Y_k=\{y_{ik}\}_{i=1}^N$ represent the pseudo-labels and true labels over all labeled examples for class $k$. There are many choices for the metric function, such as $F_{\beta}$ score, $precision$ and $recall$ \cite{chinchor1993muc}. Since $[0,1]$ is a continuous interval, it is impossible to iterate through every real value within this range. A feasible solution is to discretize this interval with a small step size. When the step size is small enough, it can be approximated that all real values with the interval have thoroughly explored. In Appendix \textcolor{red}{A}, we provide an illustration of MAT to describe its complete flow.

\begin{algorithm}[t]
	\caption{Pseudo code of the proposed algorithm.}
	\label{alg:train}
	\textbf{Input}: Labeled data $\mathcal{D}_L = \{(\x_i)^N_{i=1}, Y\}$, Unlabeled data $\mathcal{D}_U = \{\x_j\}^M_{j=1}$, backbone $f(\cdot)$, two dual-head classifiers $\{\widehat{h}^g(\cdot),\widehat{h}^l(\cdot)\}$ and $\{h^g(\cdot),h^l(\cdot)\}$, metric function $\mathcal{M}(\cdot,\cdot)$, class number $K$, small step $t$.
	\begin{algorithmic}[1] 
		\State Warm up the backbone $f(\cdot)$ and one classifier $\{\widehat{h}^g(\cdot),\widehat{h}^l(\cdot)\}$ on $\mathcal{D}_L$ with \cref{eq:lb_loss}.
		\For{each $epoch$}
		\State Input labeled data $\{\x_i\}_{i=1}^{N}$ into $f(\cdot)$ and $\{\widehat{h}^g(\cdot),\widehat{h}^l(\cdot)\}$ to get outputs $\{\hat{\q}_{i}\}_{i=1}^{N}$.
		\For{$\forall k \in [K], \tau_k=0$ to $1$ by $t$}
		\State Pseudo-label $\mathcal{D}_L$ in class $k$ by $\tau_k$, $\hat{Y}_k=\{\hat{y}_{ik}\}^N_{i=1}=\{\mathbb{I}(\hat{q}_{ik}\geq\tau_k)\}^N_{i=1}.$
		\State Select the $\tau_k$ which achieves the highest $\mathcal{M}(\hat{Y}_k, Y_k)$ as $\tau^{\star}_k$ (\cref{eq:taok_star}).
		\EndFor
		\State Pseudo-label $\mathcal{D}_U$ with \cref{eq:pseudo_label}, then train $f(\cdot)$, $\{\widehat{h}^g(\cdot),\widehat{h}^l(\cdot)\}$ and $\{h^g(\cdot),h^l(\cdot)\}$ \phantom{em} on $\mathcal{D}_L$ and $\mathcal{D}_U$ together using the D2L framework as shown in \cref{fig:framework}.
		\EndFor
	\end{algorithmic}
\end{algorithm}

Thus, our comprehensive approach combines the D2L framework and the MAT strategy, effectively enhancing the quality of pseudo-labels. For ease of understanding, we provide the algorithmic process in \Cref{alg:train}.

\section{Experiment}
\label{sec:experiment}

In this section, we first perform experiments to show the effectiveness of the proposed method; then, we conduct ablation studies to validate each component.

\subsection{Setup}
\paragraph{Datasets.}
We conduct experiments on three benchmark datasets for multi-label classification, including PASCAL VOC 2012 (VOC) \cite{EveringhamGWWZ10}, MS-COCO 2014 (COCO) \cite{LinMBHPRDZ14} and NUS-WIDE (NUS) \cite{ChuaTHLLZ09}. The detail characteristics of three benchmark datasets are reported in Appendix \textcolor{red}{B}. For the convenience and fairness of comparison, we reproduce the consistent setting with the code shared by \cite{xie2023class}. For each dataset, we randomly sample a proportion $p$ of examples from the training set that are fully annotated, and the others are unlabeled. In our experiments, we set $p=\{0.01,0.05,0.10,0.15,0.20\}$.

\paragraph{Implementation Details.}
Following \cite{xie2023class}, we use a ResNet-50 \cite{he2016deep} pre-trained on ImageNet \cite{deng2009imagenet} as the backbone and warm up the model with labeled examples for 12 epochs before training. Every input image is resized into the resolution of $224\times224$, and then is processed not only in its original form but also cropped into uniformly sized patches. In our experiments, we crop an image into a grid-like pattern with $2\times2$, $3\times3$, or $4\times4$ patches, and if not specifically indicated, each image is cropped into $2\times2$ patches by default. Both the uncropped images and the cropped patches are performed two types of data augmentation: a weak augmentation (only containing random horizontal flipping) and a strong augmentation (containing Cutout \cite{devries2017improved} and RandAugment \cite{cubuk2020randaugment}). 
In training stage, for labeled data, the predictions on strongly-augmented data and true labels are used to calculate losses, and predictions on weakly-augmented data are used to update the threshold for every class; for unlabeled data, the predictions on weakly-augmented data are used to generate pseudo-labels. In experiments, the model is trained for 40 epochs with the early stopping. We use the AdamW optimizer \cite{LoshchilovH19} with the weight decay of 1e-4, and the one cycle scheduler \cite{devries2017improved} to adjust the learning rate at the maximum of 1e-4. The batch size is set as 32 for VOC, and 64 for COCO and NUS. For MAT strategy, we use $F_{\beta}$ score as the evaluation metric, where $\beta=0.5$. In ablation experiments, we study the influence of different evaluation metrics on the performance. 
Generally, we apply exponential moving average (EMA) to the model parameters $\theta$ with a decay of 0.9997. For all methods, we use ASL \cite{ridnik2021asymmetric} as the base loss function, since it shows superiority to BCE loss, and the parameters for ASL are kept consistent with the original paper. We perform all experiments on GeForce RTX 3090 GPUs and set the random seed to 1 (the results with standard deviations can be found in Appendix \textcolor{red}{D}). Similar to the previous works \cite{xie2023class,ColeALPMJ21}, we use mean average precision (mAP) as the evaluation and report mAP on the test set. 

\begin{table}[t]
	\caption{Mean average precision (mAP \%) of each compared method on three datasets. Bold represents the highest mAP. LL-* and Top-* select the best-performing method from their respective categories. The detailed method descriptions can be found in \ref{para:methods}.}
	\label{tab:results_voc_coco_nus}
	\centering
	\begin{tabularx}{\textwidth}{l@{\hspace{5pt}}|@{\hspace{5pt}}XXXXXXXXXXX}
		\multicolumn{12}{c}{Results on VOC.} \\ 
		\toprule
		\multicolumn{1}{l}{Method} & \hspace{1pt}BCE   & \hspace{1pt}ASL   & LL-*  & \hspace{1pt}PLC   & Top-* & \hspace{2pt}IAT   & \hspace{-3pt}ADSH  & \hspace{2.5pt}FM    & \hspace{-4pt}DRML  & \hspace{1pt}CAP   & \hspace{1pt}Ours  \\
		\midrule
		$p=0.01$ & 16.71 & 34.81 & 36.01 & 43.91 & 38.61 & 34.39 & 45.06 & 44.98 & 38.90 & 41.28 & \textbf{49.09} \\
		$p=0.05$ & 67.95 & 71.46 & 75.79 & 74.49 & 75.77 & 73.24 & 75.37 & 75.11 & 61.77 & 76.16 & \textbf{79.26} \\
		$p=0.10$ & 75.35 & 78.00 & 81.04 & 80.35 & 80.78 & 80.27 & 80.34 & 80.66 & 71.01 & 82.16 & \textbf{84.06} \\
		$p=0.15$ & 78.19 & 79.69 & 82.36 & 82.35 & 82.65 & 82.39 & 82.80 & 82.63 & 72.98 & 83.48 & \textbf{86.25} \\
		$p=0.20$ & 79.38 & 80.77 & 83.68 & 83.39 & 83.72 & 83.55 & 83.93 & 83.60 & 74.49 & 84.41 & \textbf{87.16} \\
		\bottomrule
		\multicolumn{12}{c}{\ } \\ 
	\end{tabularx}
	\begin{tabularx}{\textwidth}{l@{\hspace{5pt}}|@{\hspace{5pt}}XXXXXXXXXXX}
		\multicolumn{12}{c}{Results on COCO.}\\
		\toprule
		\multicolumn{1}{l}{Method} & \hspace{1pt}BCE   & \hspace{1pt}ASL   & LL-*  & \hspace{1pt}PLC   & Top-* & \hspace{2pt}IAT   & \hspace{-3pt}ADSH  & \hspace{2.5pt}FM    & \hspace{-4pt}DRML  & \hspace{1pt}CAP   & \hspace{1pt}Ours  \\
		\midrule
		$p=0.01$ & 44.11 & 44.87 & 45.36 & 48.95 & 48.40 & 46.41 & 47.93 & 47.10 & 39.12 & 52.40 & \textbf{56.59} \\
		$p=0.05$ & 58.90 & 59.12 & 59.33 & 59.85 & 58.25 & 60.34 & 60.75 & 59.94 & 53.60 & 62.43 & \textbf{69.30} \\
		$p=0.10$ & 63.75 & 63.82 & 64.25 & 65.03 & 63.52 & 65.54 & 65.37 & 64.46 & 57.06 & 67.36 & \textbf{73.06} \\
		$p=0.15$ & 65.91 & 66.10 & 66.69 & 67.62 & 66.11 & 67.88 & 67.70 & 66.79 & 58.53 & 69.11 & \textbf{74.63} \\
		$p=0.20$ & 67.33 & 67.51 & 68.12 & 69.14 & 67.49 & 69.25 & 69.01 & 68.04 & 59.24 & 70.41 & \textbf{75.70} \\
		\bottomrule
		\multicolumn{12}{c}{\ } \\ 
	\end{tabularx}
	\begin{tabularx}{\textwidth}{l@{\hspace{5pt}}|@{\hspace{5pt}}XXXXXXXXXXX}
		\multicolumn{12}{c}{Results on NUS.} \\
		\toprule
		\multicolumn{1}{l}{Method} & \hspace{1pt}BCE   & \hspace{1pt}ASL   & LL-*  & \hspace{1pt}PLC   & Top-* & \hspace{2pt}IAT   & \hspace{-3pt}ADSH  & \hspace{2.5pt}FM    & \hspace{-4pt}DRML  & \hspace{1pt}CAP   & \hspace{1pt}Ours  \\
		\midrule
		$p=0.01$ & 29.58 & 30.51 & 20.70 & 33.59 & 26.84 & 26.28 & 33.13 & 32.10 & 17.40 & 24.75 & \textbf{38.09} \\
		$p=0.05$ & 41.09 & 42.87 & 40.20 & 43.55 & 40.99 & 42.58 & 43.94 & 43.12 & 30.61 & 44.82 & \textbf{46.86} \\
		$p=0.10$ & 45.39 & 46.50 & 44.95 & 47.51 & 45.07 & 46.60 & 47.28 & 46.65 & 35.09 & 48.24 & \textbf{50.25} \\
		$p=0.15$ & 47.30 & 48.42 & 47.32 & 49.75 & 47.43 & 48.76 & 49.22 & 48.74 & 37.91 & 49.90 & \textbf{51.61} \\
		$p=0.20$ & 48.36 & 49.65 & 48.31 & 50.71 & 48.49 & 49.62 & 49.93 & 49.59 & 39.98 & 51.06 & \textbf{52.64} \\
		\bottomrule
	\end{tabularx}
\end{table}

\paragraph{Compared Methods.}
\label{para:methods}
In this paper, the proposed method is compared against many well-established approaches, which are roughly categorized into five groups: 
\begin{itemize}
	\item Two baseline methods: \textbf{BCE} and \textbf{ASL} \cite{ridnik2021asymmetric}. We train these two methods on labeled data.
	
	\item Four MLLML methods: \textbf{LL-R}, \textbf{LL-Ct}, \textbf{LL-Cp} \cite{kim2022large} and \textbf{PLC} \cite{xie2022label}. We use \textbf{LL-*} to denote the method with the highest mAP among \textbf{LL-R}, \textbf{LL-Ct} and \textbf{LL-Cp}. 
	
	\item Three instance-aware pseudo-labeling methods: \textbf{Top-1}, \textbf{Top-$k$} and \textbf{IAT} \cite{xie2023class}. We use \textbf{Top-*} to denote the method with the highest mAP among \textbf{Top-1} and \textbf{Top-$k$}.

	\item Two SSL methods: \textbf{ADSH} \cite{guo2022class} and \textbf{FM} \cite{wang2023freematch}. Both of the aforementioned methods utilize the dynamic threshold, and we replace the $softmax(\cdot)$ activation function in them with $sigmoid(\cdot)$ to fit into multi-label datasets.

	\item Two SSMLL methods: \textbf{DRML} \cite{wang2020dual} and \textbf{CAP} \cite{xie2023class}. They currently represent the state-of-the-art SSMLL methods based on deep models. To ensure fairness, all the methods mentioned above use the same training setting as our method.
\end{itemize}



\subsection{Comparison Results}\label{sec:results}
In \Cref{tab:results_voc_coco_nus}, we report comparison results in terms of mAP on three datasets, VOC, COCO and NUS. To ensure fairness, all the methods mentioned above use the same training setting as our method. The results show that: 
1) Our approach outperforms the comparative methods on all datasets with significant improvements, particularly on the COCO dataset. Specifically, in terms of mAP, the proposed method improves upon the state-of-the-art performance by 4.19\% -- 6.87\% on COCO, 1.58\% -- 4.50\% on NUS and 1.90\% -- 4.03\% on VOC. These results demonstrate the effectiveness of our method.
2) The performance of DRML is not satisfactory, even lower than the baseline. One possible reason is that DRML only utilizes the deep model to extract image features, and during training, it only fine-tunes the linear classifier layers, without training the entire deep model end-to-end.
3) CAP generally outperforms other methods except for our approach. However, it fails at $p=0.01$ on NUS dataset. This indicates that merely estimating the class proportions is insufficient, making it challenging to achieve favorable pseudo-labeling performance when the quality of model prediction is low.
4) Compared to methods that focus solely on threshold strategies, our approach demonstrates a clear advantage, affirming the importance of model output in pseudo-labeling performance.
In addition to the results shown in \Cref{tab:results_voc_coco_nus}, we present the results of additional metrics in Appendix \textcolor{red}{F}, including evaluation metrics: average per-\textbf{C}lass \textbf{F1} (CF1) score and \textbf{O}verall \textbf{F1} (OF1) score.

\begin{figure}[t]
	\centering

	\hfill
	\begin{subfigure}{0.24\linewidth}
		\includegraphics[width=0.99\linewidth]{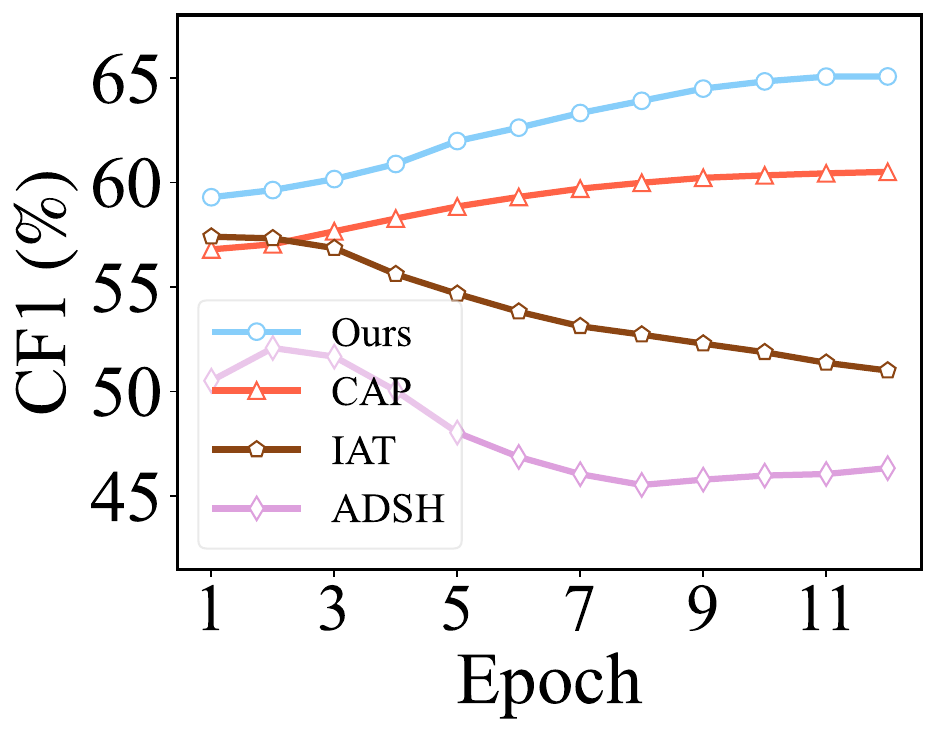}
		\caption{$p=0.05$.}
	\end{subfigure}
	\hfill
	\begin{subfigure}{0.24\linewidth}
		\includegraphics[width=0.99\linewidth]{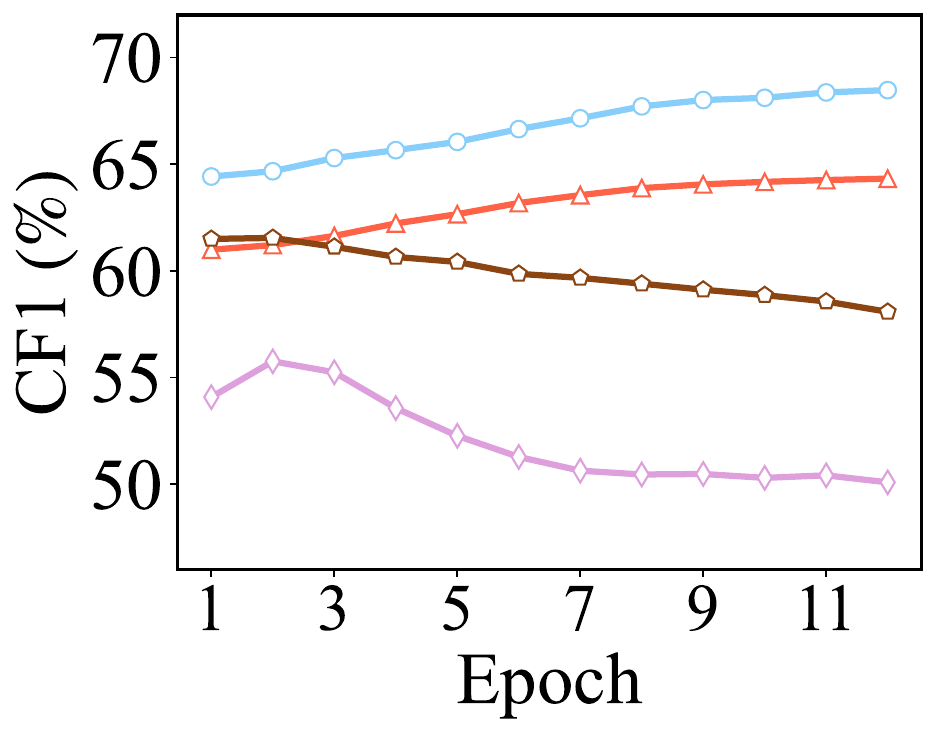}
		\caption{$p=0.1$.}
	\end{subfigure}
	\hfill
	\begin{subfigure}{0.24\linewidth}
		\includegraphics[width=0.99\linewidth]{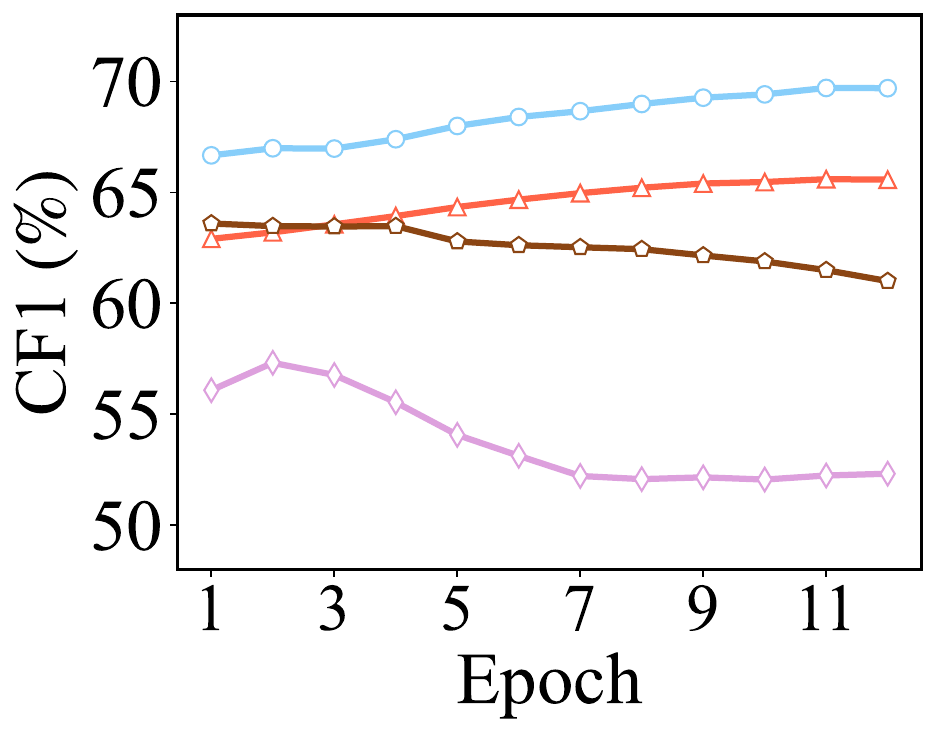}
		\caption{$p=0.15$.}
	\end{subfigure}
	\hfill
	\begin{subfigure}{0.24\linewidth}
		\includegraphics[width=0.99\linewidth]{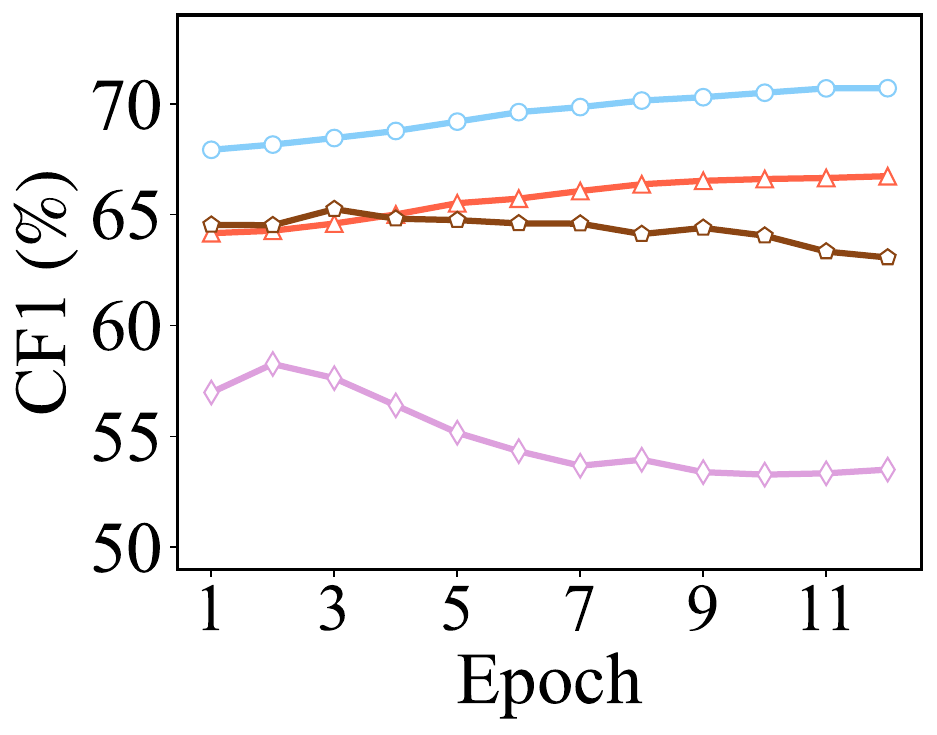}
		\caption{$p=0.2$.}
	\end{subfigure}

	\caption{The performance of pseudo-labeling during training stage on COCO.}
	\label{fig:pseudo_labeling}
\end{figure}

\subsection{Pseudo-Labeling Performance}
We perform experiments to examine the quality of pseudo-labels generated by different methods. We use CF1 score as the evaluation metric (detailed information can be found in Appendix \textcolor{red}{F}), and report results in \Cref{fig:pseudo_labeling}. Due to the page limit, the results on VOC and NUS can be found in Appendix \textcolor{red}{C}. All results were recorded after the warm-up phase. From the figures, we can see: 1) The pseudo-labeling performance of our method is always better than other methods, even compared to CAP, the state-of-the-art method in SSMLL. 2) IAT keeps a decreasing trend since introducing unlabeled examples for training, which confirms that the fixed thresholds are not proper for SSMLL. 3) ADSH consistently remains lower than the other methods. This is not surprising, given that its design is tailored for single-label scenarios. 4) From the beginning, our method has outperformed CAP because the proposed D2L framework significantly enhances the quality of model predictions. These results validate that our method can achieve better pseudo-labeling performance than the comparative methods.




\begin{table}
	\caption{Mean average precision (mAP \%) of the baseline incorporated with different components, on datasets VOC and COCO. The baseline here indicates the method CAP (the results of the first row, without any components).}
	\label{tab:results_ablation}
	\centering
	\begin{tabular}{c c c @{\hspace{5pt}}|@{\hspace{5pt}} c c c c @{\hspace{5pt}}|@{\hspace{5pt}} c c c c}
		\toprule
		\multirow{2}*{MAT} & \multicolumn{2}{c}{D2L} & \multicolumn{4}{c}{VOC} & \multicolumn{4}{c}{COCO} \\
		\cmidrule(lr){2-3}\cmidrule(lr){4-11} & CDD & GUD   & $p$=$0.05$ & $p$=$0.10$ & $p$=$0.15$ & $p$=$0.20$ & $p$=$0.05$ & $p$=$0.10$ & $p$=$0.15$ & $p$=$0.20$ \\ \midrule
		&		&		&	76.16	&	82.16	&	83.48	&	84.41	&	62.43	&	67.36	&	69.11	&	70.41	\\	\midrule
		\checkmark	&		&		&	76.87	&	82.59	&	84.29	&	85.16	&	65.03	&	68.87	&	70.54	&	71.54	\\	
		\checkmark	&	\checkmark	&		&	77.11	&	83.48	&	85.72	&	86.55	&	66.07	&	70.72	&	72.92	&	74.26	\\	
		\checkmark	&	\checkmark	&	\checkmark	&	79.26	&	84.06	&	86.25	&	87.16	&	69.30	&	73.06	&	74.63	&	75.70	\\	
		\bottomrule
	\end{tabular}
\end{table}

\subsection{Ablation Studies}

\begin{figure}[t]
	\centering
	\begin{subfigure}{0.24\linewidth}
		\includegraphics[width=0.99\linewidth]{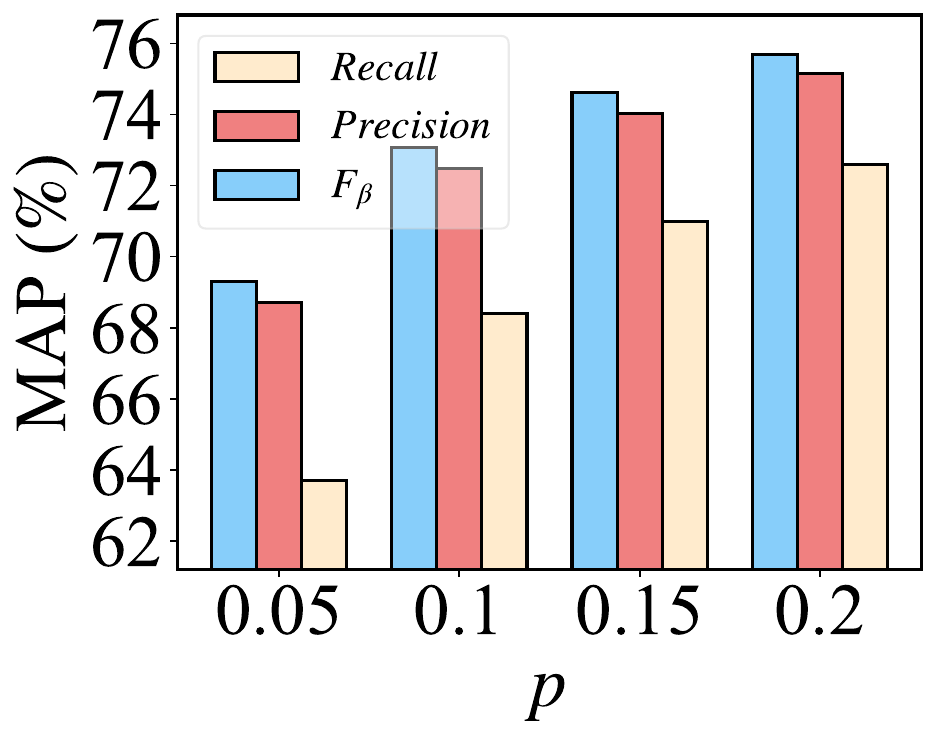}
		\caption{The ablation study on metric $\mathcal{M}(\cdot,\cdot)$.}
		\label{fig:coco_mat}
	\end{subfigure}
	\hfill
	\begin{subfigure}{0.24\linewidth}
		\includegraphics[width=0.99\linewidth]{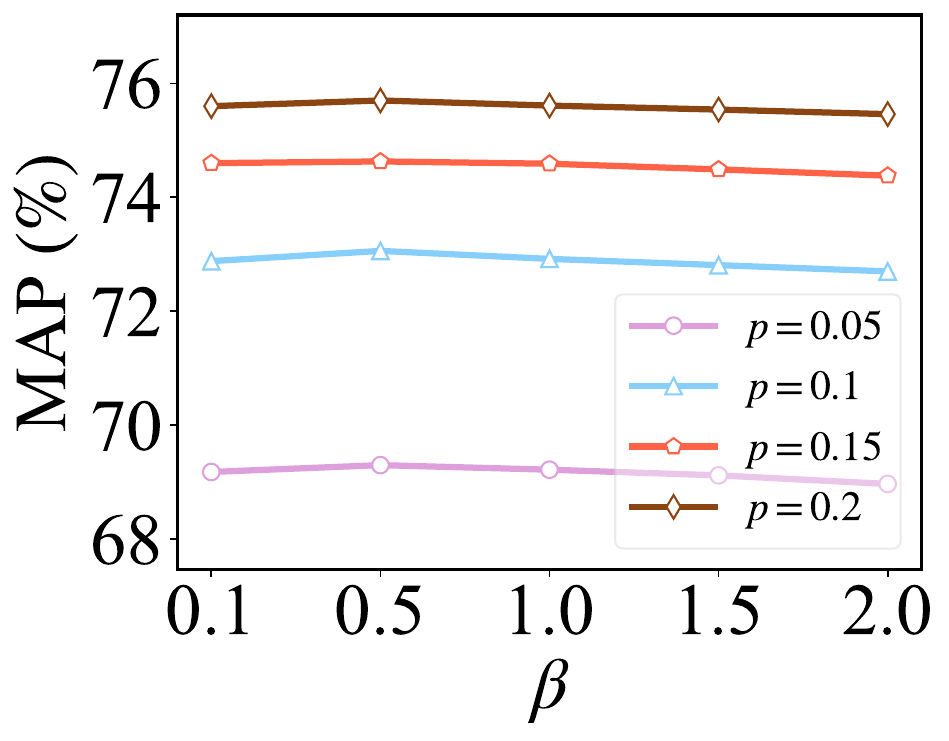}
		\caption{The ablation study on $\beta$ in metric $F_\beta$.}
		\label{fig:coco_fb}
	\end{subfigure}
	\hfill
	\begin{subfigure}{0.24\linewidth}
		\includegraphics[width=1.05\linewidth]{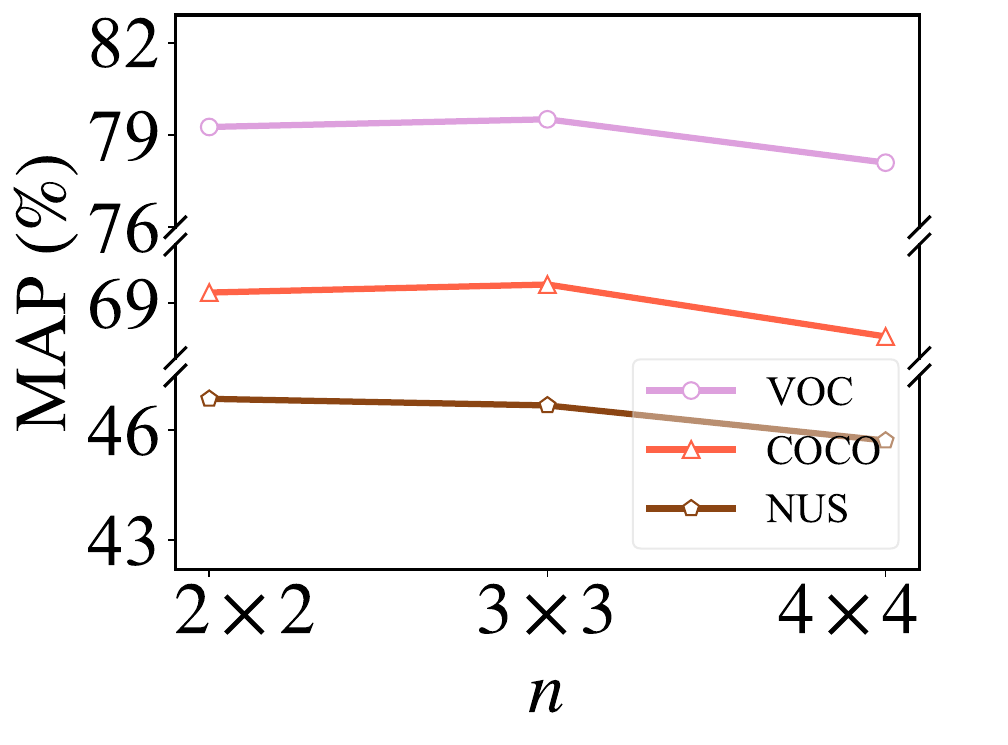}
		\caption{The ablation study on number of patches $n$.}
		\label{fig:0.05_grid}
	\end{subfigure}
	\hfill
	\begin{subfigure}{0.24\linewidth}
		\includegraphics[width=0.99\linewidth]{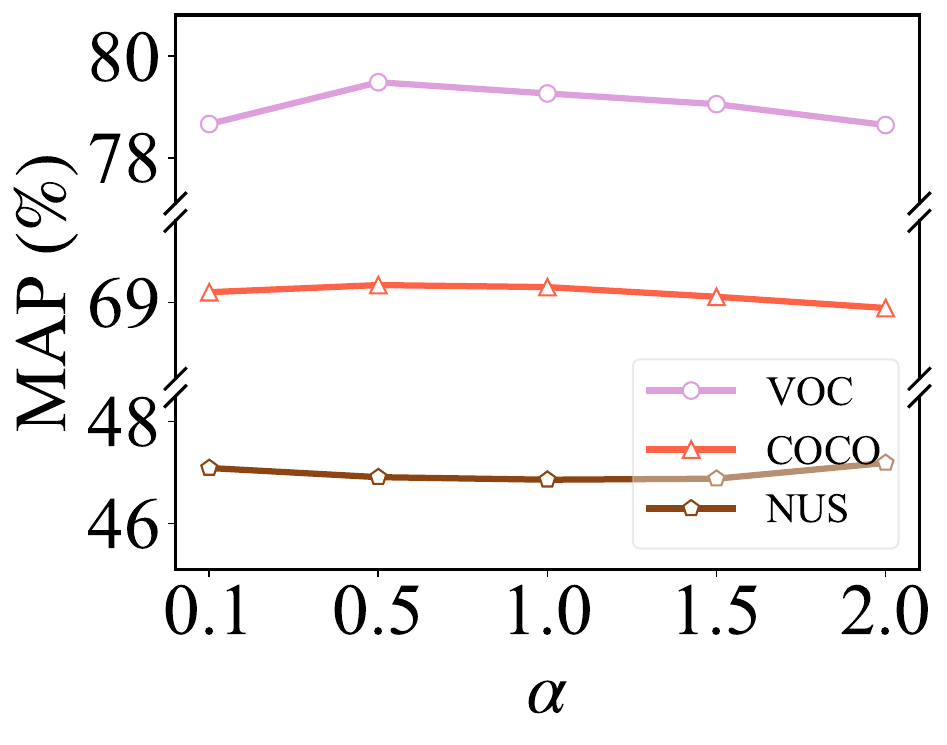}
		\caption{The ablation study on temperature $\alpha$.}
		\label{fig:0.05_tem}
	\end{subfigure}
	\caption{The analyses of parameters in D2L and MAT: (a-b) The results of various metric functions $\mathcal{M}(\cdot,\cdot)$ used in MAT and different $\beta$ values used in metric $F_\beta$, at $p=\{0.05,0.1,0.15,0.2\}$ on COCO; (c-d) The analyses of two parameters, number of patches $n$ and temperature $\alpha$ in D2L framework, at $p=0.05$ on three datasets. The parameter analyses under other settings will be presented in Appendix \textcolor{red}{E}.}
	
	\label{fig:ablation_mat}
\end{figure}

\paragraph{The Study on D2L and MAT.}
To further investigate how the proposed method improves the performance of SSMLL, we conduct a series of ablation experiments on VOC and COCO datasets at four different labeled proportions $p=\{0.05,0.10,0.15,0.20\}$. Since our method consists of two main components, D2L (including CDD and GUD, which indicate `Correlative and Discriminative features Decoupling' and `Generation and Utilization of pseudo-labels Decoupling', respectively) and MAT, we incrementally introduce these components to demonstrate the improvement of each part. The results are reported in \Cref{tab:results_ablation}. It is noteworthy that the method in the first row, which does not contain any of these components, is CAP (mentioned in \Cref{para:methods}). This method adopt a standard classifier and use the class-distribution-aware thresholding method. We consider it as the baseline in ablation experiments. 
Firstly, we abandon the CAP in favor of the MAT, which resulted in a 0.43\% -- 0.81\% improvement on VOC and a 1.13\% -- 2.60\% improvement on COCO. The MAT strategy shows more significant improvements as the number of labeled examples decreases, which indirectly indicates that MAT can make better use of unlabeled data compared to CAP strategy, even in scenarios where labeled data is scarce.
Secondly, we replace the standard classifier with a dual-head classifier $\{h^g(\cdot),h^l(\cdot)\}$ and then crop images into patches for training. Note that during this period the classifier is shared by both labeled and unlabeled data, and the strategy of pseudo-labeling is MAT. It can be observed that the performance has improved by 0.24\% -- 1.43\% on VOC and 1.04\% -- 2.72\% on COCO. This validates the effectiveness of dual-head classifier on decoupling the learning of correlative and discriminative features. 
Finally, to mitigate the influence of unlabeled data on the true distribution learning, we introduce another dual-head classifier $\{\widehat{h}^g(\cdot),\widehat{h}^l(\cdot)\}$ dedicated to serving the learning of labeled data, and $\{h^g(\cdot),h^l(\cdot)\}$ is responsible only for learning from unlabeled data. After doing that, the performance has improved by 0.53\% -- 2.15\% on VOC and 1.44\% -- 3.23\% on COCO, demonstrating the importance of segregating the learning of generation and utilization of pseudo-labels. All these results convincingly validate that each component of the proposed method makes positive contribution to the final performance.
\paragraph{The Study on Metric Function.}
We study the influence of different metric functions on the final performance. \Cref{fig:coco_mat} illustrates the performance as the metric function $\mathcal{M}$ changes in $\{F_\beta,Precision,Recall\}$, where $\beta=0.5$. Moreover, we conduct the further research on the choice of $\beta$ values in \Cref{fig:coco_fb}. We observe that $F_\beta$ metric performs better than the other two metrics on COCO. This is because $F_\beta=((1+\beta^2) * P * R) /(\beta * P + R)$ combines the other two metrics $Precision$ ($P$) and $Recall$ ($R$) to provide a more comprehensive performance evaluation. For different values of $\beta$, we find that MAT is insensitive to this parameter. So, we can set this value safely in practice.

\paragraph{Parameter Sensitivity Analyses.}
In \Cref{fig:0.05_grid,fig:0.05_tem}, we show that the performance changes as parameters $n$ and $\alpha$ are varied within the range $\{2\times2, 3\times3, 4\times4\}$ and $\{0.1,0.5,1.0,1.5,2.0\}$. As shown in \Cref{fig:0.05_grid}, the performance degrades with the increase of $n$. This is mainly because when an image is cropped into too many small patches, most of these patches contain semantically irrelevant backgrounds, and a complete semantic object may also be distributed across different patches. The phenomenon makes the model hard to recognize interest objects, thus leading to performance degradation. Taking both computational cost and model performance into consideration, we uniformly adopt $n=2\times2$ in our experiments. Regarding the temperature $\alpha$ shown in \Cref{fig:0.05_tem}, the performance of our method is generally insensitive to $\alpha$ on COCO and NUS, while on VOC, we generally find that choosing a too large or small temperature value would degrade the model performance. Considering both universality and simplicity, we choose $\alpha=1.0$ for all experiments. 



\begin{figure}
	\centering
	
	\includegraphics[width=0.85\linewidth]{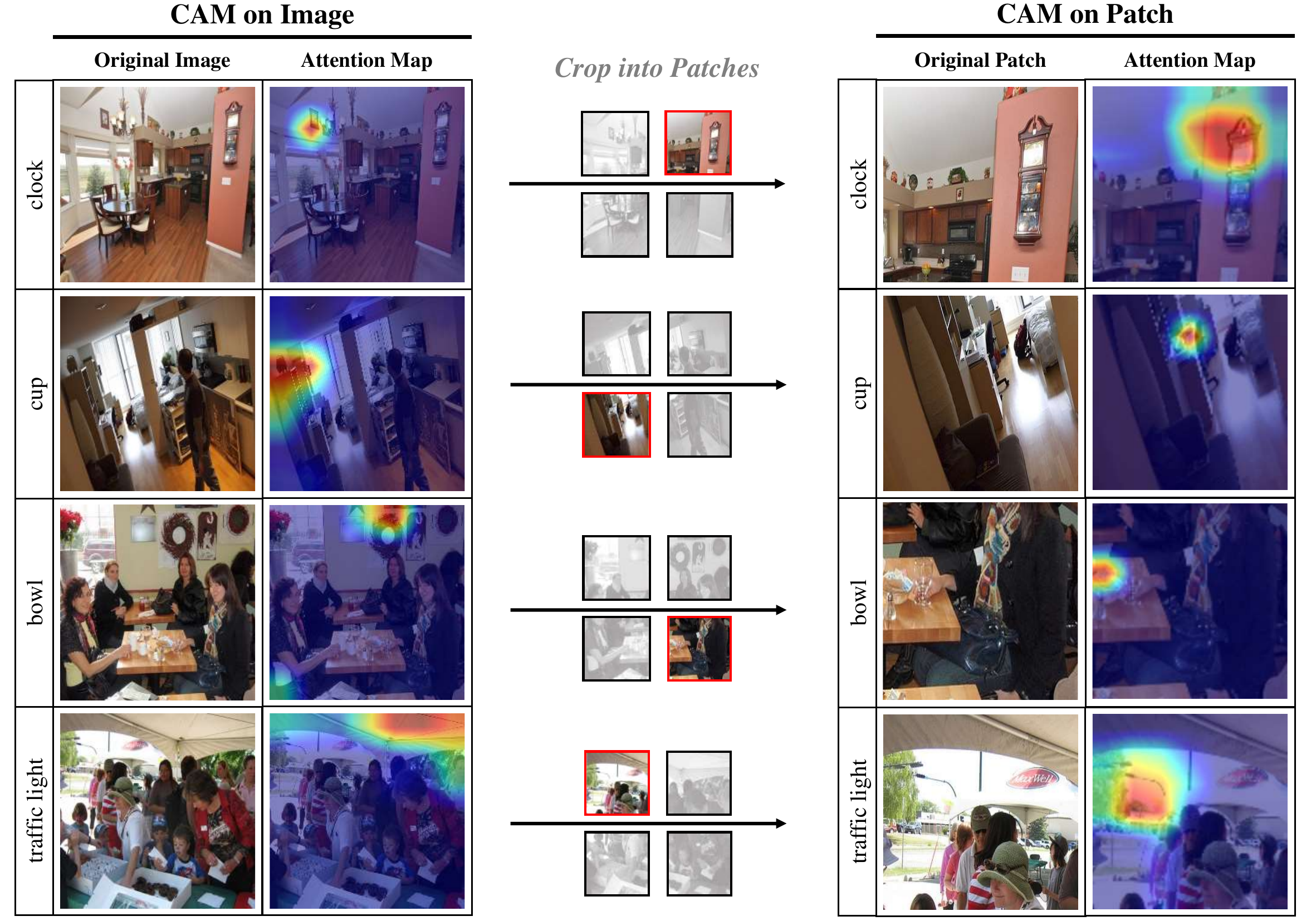}
	
	\caption{Visualization of attention maps on COCO. Each patch is cropped from the original image starting from the beginning of a row. The class label attached in front of every original image or cropped patch is activated in the attention map.}
	\label{fig:case}
\end{figure}

\subsection{Case Study}
To disclose the mechanism behind the effectiveness of D2L for producing high-quality predictions, \Cref{fig:case} visualizes some attention maps on COCO. For every original image (left) and cropped patch (right), we show the attention map of a given class label. From these figures, we observe that the model typically makes mistakes when identifying indistinguishable objects in the whole image, \eg, ambiguous objects or tiny objects, while it is able to correct these mistakes by precisely identifying these objects in the patches. For example, in the first image, the model mislabels a square painting as a \textit{clock} in the original image, while it corrects this error by precisely identifying the object in the corresponding patch. These results convincingly verify that our proposed D2L framework can significantly enhance the ability of model for identifying indistinguishable targets, leading to high-quality pseudo-labels.


%



\section{Conclusion}
\label{sec:conclusion}

This paper studies the problem of semi-supervised multi-label learning (SSMLL). We propose a dual-perspective method to boost the pseudo-labeling performance. To improve the quality of model predictions, we develop the D2L method to perform correlative/discriminative features decoupling, which enhances the feature learning, and generation/utilization of pseudo-labels decoupling, which avoids the accumulation of incorrect pseudo-labels. 
To achieve proper thresholding, we propose to estimate the thresholds in a metric-adaptive manner, with the goal of maximizing the pseudo-labeling measurement on labeled data.
Extensive experimental results on multiple benchmark datasets validate that the proposed method can achieve the state-of-the-art performance. In the future, we will improve the performance of SSMLL by incorporating the knowledge of pre-trained vision-language models.

\section*{Acknowledgements}
Sheng-Jun Huang was supported by Natural Science Foundation of Jiangsu Province of China (BK20222012, BK20211517), the National Key R\&D Program of China (2020AAA0107000), and NSFC (62222605). 
Masashi Sugiyama was supported by JST CREST Grant Number JPMJCR18A2 and a grant from Apple, Inc. Any views, opinions, findings, and conclusions or recommendations expressed in this material are those of the authors and should not be interpreted as reflecting the views, policies or position, either expressed or implied, of Apple Inc.

%
%
\bibliographystyle{splncs04}
\bibliography{main}

\begin{thebibliography}{10}
\providecommand{\url}[1]{\texttt{#1}}
\providecommand{\urlprefix}{URL }
\providecommand{\doi}[1]{https://doi.org/#1}

\bibitem{ben2022multi}
Ben-Baruch, E., Ridnik, T., Friedman, I., Ben-Cohen, A., Zamir, N., Noy, A.,
  Zelnik-Manor, L.: Multi-label classification with partial annotations using
  class-aware selective loss. In: Proceedings of the IEEE/CVF Conference on
  Computer Vision and Pattern Recognition. pp. 4764--4772 (2022)

\bibitem{chen2022debiased}
Chen, B., Jiang, J., Wang, X., Wan, P., Wang, J., Long, M.: Debiased
  self-training for semi-supervised learning. Advances in Neural Information
  Processing Systems  \textbf{35},  32424--32437 (2022)

\bibitem{chen2008semi}
Chen, G., Song, Y., Wang, F., Zhang, C.: Semi-supervised multi-label learning
  by solving a sylvester equation. In: Proceedings of the 2008 SIAM
  international conference on data mining. pp. 410--419. SIAM (2008)

\bibitem{chen2019learning}
Chen, T., Xu, M., Hui, X., Wu, H., Lin, L.: Learning semantic-specific graph
  representation for multi-label image recognition. In: Proceedings of the
  IEEE/CVF international conference on computer vision. pp. 522--531 (2019)

\bibitem{chen2019multi}
Chen, Z.M., Wei, X.S., Wang, P., Guo, Y.: Multi-label image recognition with
  graph convolutional networks. In: Proceedings of the IEEE/CVF conference on
  computer vision and pattern recognition. pp. 5177--5186 (2019)

\bibitem{chinchor1993muc}
Chinchor, N., Sundheim, B.M.: Muc-5 evaluation metrics. In: Fifth Message
  Understanding Conference (MUC-5): Proceedings of a Conference Held in
  Baltimore, Maryland, August 25-27, 1993 (1993)

\bibitem{ChuaTHLLZ09}
Chua, T., Tang, J., Hong, R., Li, H., Luo, Z., Zheng, Y.: {NUS-WIDE:} a
  real-world web image database from national university of singapore. In:
  Marchand{-}Maillet, S., Kompatsiaris, Y. (eds.) Proceedings of the 8th {ACM}
  International Conference on Image and Video Retrieval (2009)

\bibitem{ColeALPMJ21}
Cole, E., Aodha, O.M., Lorieul, T., Perona, P., Morris, D., Jojic, N.:
  Multi-label learning from single positive labels. In: {IEEE} Conference on
  Computer Vision and Pattern Recognition. pp. 933--942 (2021)

\bibitem{cubuk2020randaugment}
Cubuk, E.D., Zoph, B., Shlens, J., Le, Q.V.: Randaugment: Practical automated
  data augmentation with a reduced search space. In: Proceedings of the
  IEEE/CVF conference on computer vision and pattern recognition workshops. pp.
  702--703 (2020)

\bibitem{deng2009imagenet}
Deng, J., Dong, W., Socher, R., Li, L.J., Li, K., Fei-Fei, L.: Imagenet: A
  large-scale hierarchical image database. In: 2009 IEEE conference on computer
  vision and pattern recognition. pp. 248--255. Ieee (2009)

\bibitem{devries2017improved}
DeVries, T., Taylor, G.W.: Improved regularization of convolutional neural
  networks with cutout. arXiv preprint arXiv:1708.04552  (2017)

\bibitem{dosovitskiy2020image}
Dosovitskiy, A., Beyer, L., Kolesnikov, A., Weissenborn, D., Zhai, X.,
  Unterthiner, T., Dehghani, M., Minderer, M., Heigold, G., Gelly, S., et~al.:
  An image is worth 16x16 words: Transformers for image recognition at scale.
  arXiv preprint arXiv:2010.11929  (2020)

\bibitem{durand2019learning}
Durand, T., Mehrasa, N., Mori, G.: Learning a deep convnet for multi-label
  classification with partial labels. In: Proceedings of the IEEE/CVF
  conference on computer vision and pattern recognition. pp. 647--657 (2019)

\bibitem{EveringhamGWWZ10}
Everingham, M., Gool, L.V., Williams, C.K.I., Winn, J.M., Zisserman, A.: The
  pascal visual object classes {(VOC)} challenge. International Journal of
  Computer Vision  \textbf{88}(2),  303--338 (2010)

\bibitem{guo2021long}
Guo, H., Wang, S.: Long-tailed multi-label visual recognition by collaborative
  training on uniform and re-balanced samplings. In: Proceedings of the
  IEEE/CVF Conference on Computer Vision and Pattern Recognition. pp.
  15089--15098 (2021)

\bibitem{guo2022class}
Guo, L.Z., Li, Y.F.: Class-imbalanced semi-supervised learning with adaptive
  thresholding. In: International Conference on Machine Learning. pp.
  8082--8094. PMLR (2022)

\bibitem{guo2012semi}
Guo, Y., Schuurmans, D.: Semi-supervised multi-label classification: a
  simultaneous large-margin, subspace learning approach. In: Machine Learning
  and Knowledge Discovery in Databases: European Conference, ECML PKDD 2012,
  Bristol, UK, September 24-28, 2012. Proceedings, Part II 23. pp. 355--370.
  Springer (2012)

\bibitem{gupta2019lvis}
Gupta, A., Dollar, P., Girshick, R.: Lvis: A dataset for large vocabulary
  instance segmentation. In: Proceedings of the IEEE/CVF conference on computer
  vision and pattern recognition. pp. 5356--5364 (2019)

\bibitem{he2016deep}
He, K., Zhang, X., Ren, S., Sun, J.: Deep residual learning for image
  recognition. In: Proceedings of the IEEE conference on computer vision and
  pattern recognition. pp. 770--778 (2016)

\bibitem{hu2019weakly}
Hu, M., Han, H., Shan, S., Chen, X.: Weakly supervised image classification
  through noise regularization. In: Proceedings of the IEEE/CVF Conference on
  Computer Vision and Pattern Recognition. pp. 11517--11525 (2019)

\bibitem{huang2012multi}
Huang, S.J., Zhou, Z.H.: Multi-label learning by exploiting label correlations
  locally. In: Proceedings of the AAAI Conference on Artificial Intelligence.
  vol.~26, pp. 949--955 (2012)

\bibitem{huynh2020interactive}
Huynh, D., Elhamifar, E.: Interactive multi-label cnn learning with partial
  labels. In: Proceedings of the IEEE/CVF Conference on Computer Vision and
  Pattern Recognition. pp. 9423--9432 (2020)

\bibitem{kim2022large}
Kim, Y., Kim, J.M., Akata, Z., Lee, J.: Large loss matters in weakly supervised
  multi-label classification. In: Proceedings of the IEEE/CVF Conference on
  Computer Vision and Pattern Recognition. pp. 14156--14165 (2022)

\bibitem{kuznetsova2020open}
Kuznetsova, A., Rom, H., Alldrin, N., Uijlings, J., Krasin, I., Pont-Tuset, J.,
  Kamali, S., Popov, S., Malloci, M., Kolesnikov, A., et~al.: The open images
  dataset v4: Unified image classification, object detection, and visual
  relationship detection at scale. International Journal of Computer Vision
  \textbf{128}(7),  1956--1981 (2020)

\bibitem{lanchantin2021general}
Lanchantin, J., Wang, T., Ordonez, V., Qi, Y.: General multi-label image
  classification with transformers. In: Proceedings of the IEEE/CVF Conference
  on Computer Vision and Pattern Recognition. pp. 16478--16488 (2021)

\bibitem{lee2021abc}
Lee, H., Shin, S., Kim, H.: Abc: Auxiliary balanced classifier for
  class-imbalanced semi-supervised learning. Advances in Neural Information
  Processing Systems  \textbf{34},  7082--7094 (2021)

\bibitem{li2021online}
Li, P., Wang, H., B{\"o}hm, C., Shao, J.: Online semi-supervised multi-label
  classification with label compression and local smooth regression. In:
  Proceedings of the Twenty-Ninth International Conference on International
  Joint Conferences on Artificial Intelligence. pp. 1359--1365 (2021)

\bibitem{li2022estimating}
Li, S., Xia, X., Zhang, H., Zhan, Y., Ge, S., Liu, T.: Estimating noise
  transition matrix with label correlations for noisy multi-label learning.
  Advances in Neural Information Processing Systems  \textbf{35},  24184--24198
  (2022)

\bibitem{lin2018semantic}
Lin, J., Su, Q., Yang, P., Ma, S., Sun, X.: Semantic-unit-based dilated
  convolution for multi-label text classification. arXiv preprint
  arXiv:1808.08561  (2018)

\bibitem{LinMBHPRDZ14}
Lin, T.Y., Maire, M., Belongie, S., Hays, J., Perona, P., Ramanan, D.,
  Doll{\'a}r, P., Zitnick, C.L.: Microsoft coco: Common objects in context. In:
  European conference on computer vision. pp. 740--755 (2014)

\bibitem{liu2021emerging}
Liu, W., Wang, H., Shen, X., Tsang, I.W.: The emerging trends of multi-label
  learning. IEEE transactions on pattern analysis and machine intelligence
  \textbf{44}(11),  7955--7974 (2021)

\bibitem{liu2021swin}
Liu, Z., Lin, Y., Cao, Y., Hu, H., Wei, Y., Zhang, Z., Lin, S., Guo, B.: Swin
  transformer: Hierarchical vision transformer using shifted windows. In:
  Proceedings of the IEEE/CVF international conference on computer vision. pp.
  10012--10022 (2021)

\bibitem{LoshchilovH19}
Loshchilov, I., Hutter, F.: Decoupled weight decay regularization. In: 7th
  International Conference on Learning Representations (2019)

\bibitem{ridnik2021asymmetric}
Ridnik, T., Ben-Baruch, E., Zamir, N., Noy, A., Friedman, I., Protter, M.,
  Zelnik-Manor, L.: Asymmetric loss for multi-label classification. In:
  Proceedings of the IEEE/CVF International Conference on Computer Vision. pp.
  82--91 (2021)

\bibitem{sohn2020fixmatch}
Sohn, K., Berthelot, D., Carlini, N., Zhang, Z., Zhang, H., Raffel, C.A.,
  Cubuk, E.D., Kurakin, A., Li, C.L.: Fixmatch: Simplifying semi-supervised
  learning with consistency and confidence. Advances in neural information
  processing systems  \textbf{33},  596--608 (2020)

\bibitem{tan2017semi}
Tan, Q., Yu, Y., Yu, G., Wang, J.: Semi-supervised multi-label classification
  using incomplete label information. Neurocomputing  \textbf{260},  192--202
  (2017)

\bibitem{touvron2021training}
Touvron, H., Cord, M., Douze, M., Massa, F., Sablayrolles, A., J{\'e}gou, H.:
  Training data-efficient image transformers \& distillation through attention.
  In: International conference on machine learning. pp. 10347--10357. PMLR
  (2021)

\bibitem{wang2013dynamic}
Wang, B., Tu, Z., Tsotsos, J.K.: Dynamic label propagation for semi-supervised
  multi-class multi-label classification. In: Proceedings of the IEEE
  international conference on computer vision. pp. 425--432 (2013)

\bibitem{wang2016cnn}
Wang, J., Yang, Y., Mao, J., Huang, Z., Huang, C., Xu, W.: Cnn-rnn: A unified
  framework for multi-label image classification. In: Proceedings of the IEEE
  conference on computer vision and pattern recognition. pp. 2285--2294 (2016)

\bibitem{wang2020dual}
Wang, L., Liu, Y., Qin, C., Sun, G., Fu, Y.: Dual relation semi-supervised
  multi-label learning. In: Proceedings of the AAAI Conference on Artificial
  Intelligence. vol.~34, pp. 6227--6234 (2020)

\bibitem{wang2023imbalanced}
Wang, R., Jia, X., Wang, Q., Wu, Y., Meng, D.: Imbalanced semi-supervised
  learning with bias adaptive classifier. In: 11th International Conference on
  Learning Representations (ICLR 2023) (2023)

\bibitem{wang2023freematch}
Wang, Y., Chen, H., Heng, Q., Hou, W., Fan, Y., Wu, Z., Wang, J., Savvides, M.,
  Shinozaki, T., Raj, B., Schiele, B., Xie, X.: Freematch: Self-adaptive
  thresholding for semi-supervised learning. In: The Eleventh International
  Conference on Learning Representations (2023)

\bibitem{xie2018partial}
Xie, M.K., Huang, S.J.: Partial multi-label learning. In: Proceedings of the
  AAAI Conference on Artificial Intelligence. vol.~32 (2018)

\bibitem{xie2023class}
Xie, M.K., Xiao, J.H., Niu, G., Sugiyama, M., Huang, S.J.:
  Class-distribution-aware pseudo labeling for semi-supervised multi-label
  learning. arXiv preprint arXiv:2305.02795  (2023)

\bibitem{xie2024counterfactualreasoningmultilabelimage}
Xie, M.K., Xiao, J.H., Peng, P., Niu, G., Sugiyama, M., Huang, S.J.:
  Counterfactual reasoning for multi-label image classification via
  patching-based training (2024)

\bibitem{xie2022label}
Xie, M.K., Xiao, J., Huang, S.J.: Label-aware global consistency for
  multi-label learning with single positive labels. Advances in Neural
  Information Processing Systems  \textbf{35},  18430--18441 (2022)

\bibitem{yazici2020orderless}
Yazici, V.O., Gonzalez-Garcia, A., Ramisa, A., Twardowski, B., Weijer, J.v.d.:
  Orderless recurrent models for multi-label classification. In: Proceedings of
  the IEEE/CVF Conference on Computer Vision and Pattern Recognition. pp.
  13440--13449 (2020)

\bibitem{ye2020attention}
Ye, J., He, J., Peng, X., Wu, W., Qiao, Y.: Attention-driven dynamic graph
  convolutional network for multi-label image recognition. In: Computer
  Vision--ECCV 2020: 16th European Conference, Glasgow, UK, August 23--28,
  2020, Proceedings, Part XXI 16. pp. 649--665. Springer (2020)

\bibitem{zhan2017inductive}
Zhan, W., Zhang, M.L.: Inductive semi-supervised multi-label learning with
  co-training. In: Proceedings of the 23rd ACM SIGKDD international conference
  on knowledge discovery and data mining. pp. 1305--1314 (2017)

\bibitem{zhang2021flexmatch}
Zhang, B., Wang, Y., Hou, W., Wu, H., Wang, J., Okumura, M., Shinozaki, T.:
  Flexmatch: Boosting semi-supervised learning with curriculum pseudo labeling.
  Advances in Neural Information Processing Systems  \textbf{34},  18408--18419
  (2021)

\bibitem{zhang2021multi}
Zhang, D., Ju, X., Zhang, W., Li, J., Li, S., Zhu, Q., Zhou, G.: Multi-modal
  multi-label emotion recognition with heterogeneous hierarchical message
  passing. In: Proceedings of the AAAI conference on artificial intelligence.
  vol.~35, pp. 14338--14346 (2021)

\bibitem{zhang2020partial}
Zhang, M.L., Fang, J.P.: Partial multi-label learning via credible label
  elicitation. IEEE Transactions on Pattern Analysis and Machine Intelligence
  \textbf{43}(10),  3587--3599 (2020)

\bibitem{zhang2018binary}
Zhang, M.L., Li, Y.K., Liu, X.Y., Geng, X.: Binary relevance for multi-label
  learning: an overview. Frontiers of Computer Science  \textbf{12},  191--202
  (2018)

\bibitem{zhao2015semi}
Zhao, F., Guo, Y.: Semi-supervised multi-label learning with incomplete labels.
  In: Twenty-fourth international joint conference on artificial intelligence
  (2015)

\bibitem{zhou2012multi}
Zhou, Z.H., Zhang, M.L., Huang, S.J., Li, Y.F.: Multi-instance multi-label
  learning. Artificial Intelligence  \textbf{176}(1),  2291--2320 (2012)

\bibitem{zhu2017learning}
Zhu, F., Li, H., Ouyang, W., Yu, N., Wang, X.: Learning spatial regularization
  with image-level supervisions for multi-label image classification. In:
  Proceedings of the IEEE conference on computer vision and pattern
  recognition. pp. 5513--5522 (2017)

\bibitem{zhu2017multi}
Zhu, Y., Kwok, J.T., Zhou, Z.H.: Multi-label learning with global and local
  label correlation. IEEE Transactions on Knowledge and Data Engineering
  \textbf{30}(6),  1081--1094 (2017)

\end{thebibliography}

\clearpage
\setcounter{page}{1}

\appendix

\section{An Illustration of MAT}
\label{app:mat}
\begin{figure}[h]
	\centering
	\includegraphics[width=0.99\linewidth]{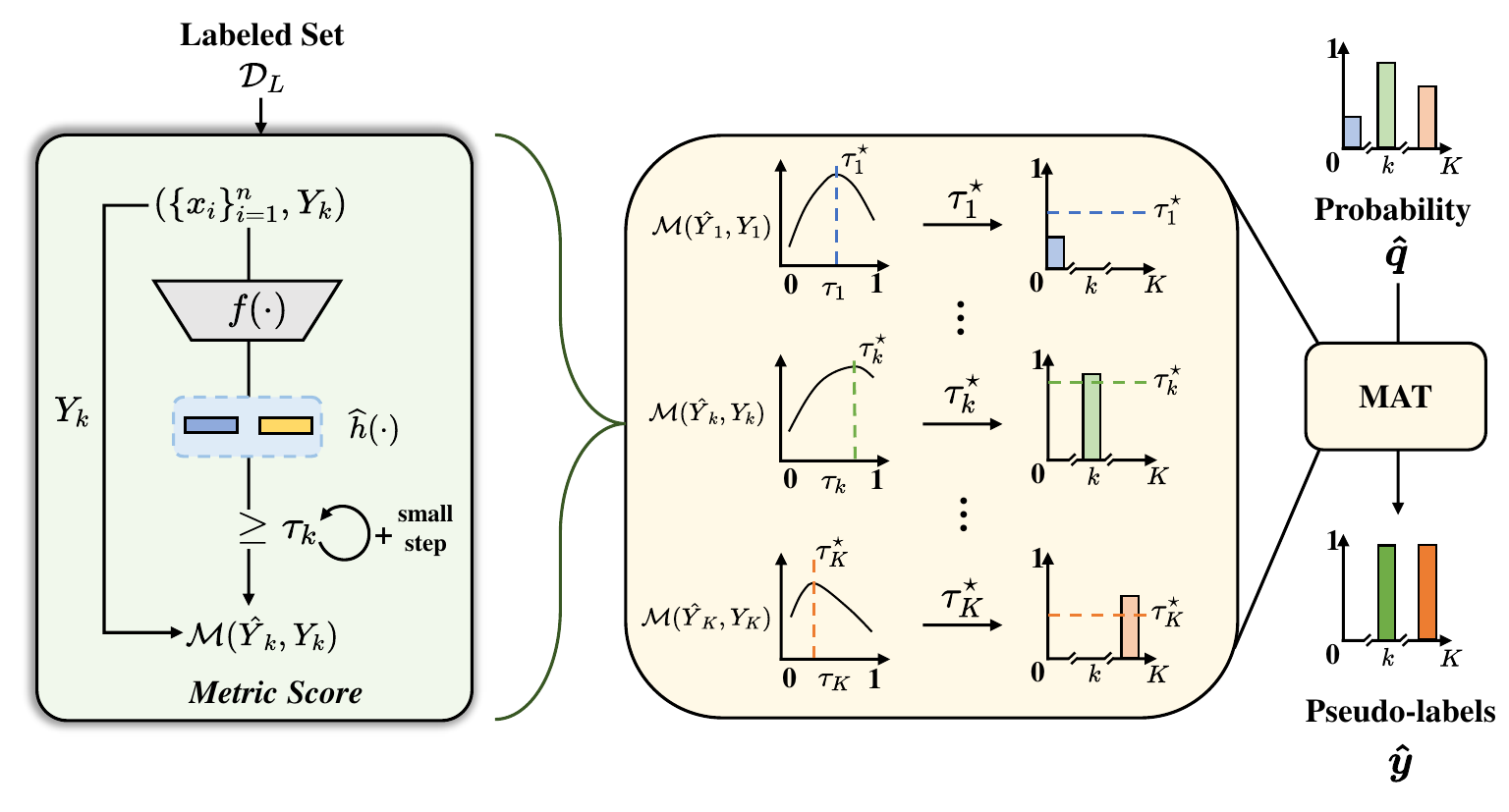}
	\caption{An illustration of MAT. By feeding instances into the model $f(\cdot)\circ\widehat{h}(\cdot)$, we obtain the predictions. By adjusting $\tau_k$, we can achieve the optimal pseudo-labeling performance  $\mathcal{M}(\hat{Y}_k, Y_k)$.}
	\label{fig:mat}
\end{figure}

\begin{table}[h]
	\centering
	\caption{The detailed characteristics of three benchmark datasets.}
	\label{tab:data}
	\begin{tabular}{lcccc}
		\toprule
		\centering{Dataset} & \# Class. & \# Train. & \# Val. &  \# Avg.\\
		\midrule
		VOC  	& 20 & 5,717	& 5,823		&  1.5	\\
		COCO    & 80 & 82,081	& 40,137	&  2.9 \\
		NUS  	& 81 & 150,000	& 60,260	&  1.9	\\
		\bottomrule
	\end{tabular}
\end{table}

\section{Details of Datasets}
\label{app:datasets}
The detail characteristics of three benchmark datasets, including PASCAL VOC 2012 (VOC) \cite{EveringhamGWWZ10}, MS-COCO 2014 (COCO) \cite{LinMBHPRDZ14} and NUS-WIDE (NUS), \cite{ChuaTHLLZ09} are reported in \Cref{tab:data}. Specifically, VOC contains 5,717 training images and 5,823 validation images for 20 classes. The average number of labels per image in VOC is 1.5. For COCO, there are 82,081 training images and 40,137 validation images for 80 classes, and the average number of labels per image is 2.9. Following \cite{xie2023class}, we split NUS as 150,000 training images and 60,260 validation images, containing 81 classes, where the average number of labels per image is 1.9. In our experiments, we report results on the above validation sets of three datasets.

\begin{figure}[t]
	\centering
	\begin{subfigure}{0.99\linewidth}
		\includegraphics[width=1.0\linewidth]{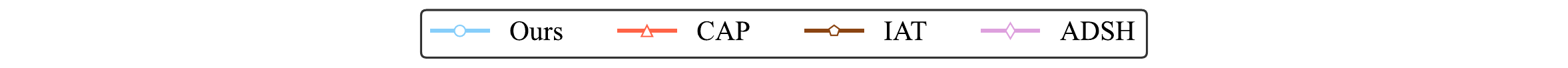}
	\end{subfigure}
	\hfill
	\begin{subfigure}{0.24\linewidth}
		\includegraphics[width=0.99\linewidth]{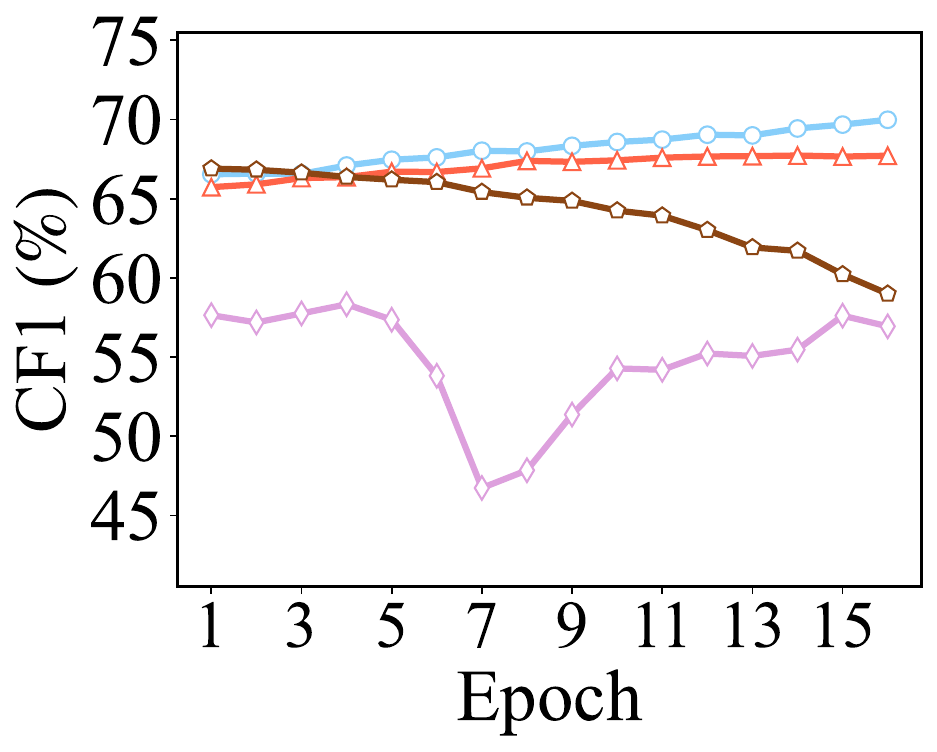}
		\caption{$p=0.05$.}
	\end{subfigure}
	\hfill
	\begin{subfigure}{0.24\linewidth}
		\includegraphics[width=0.99\linewidth]{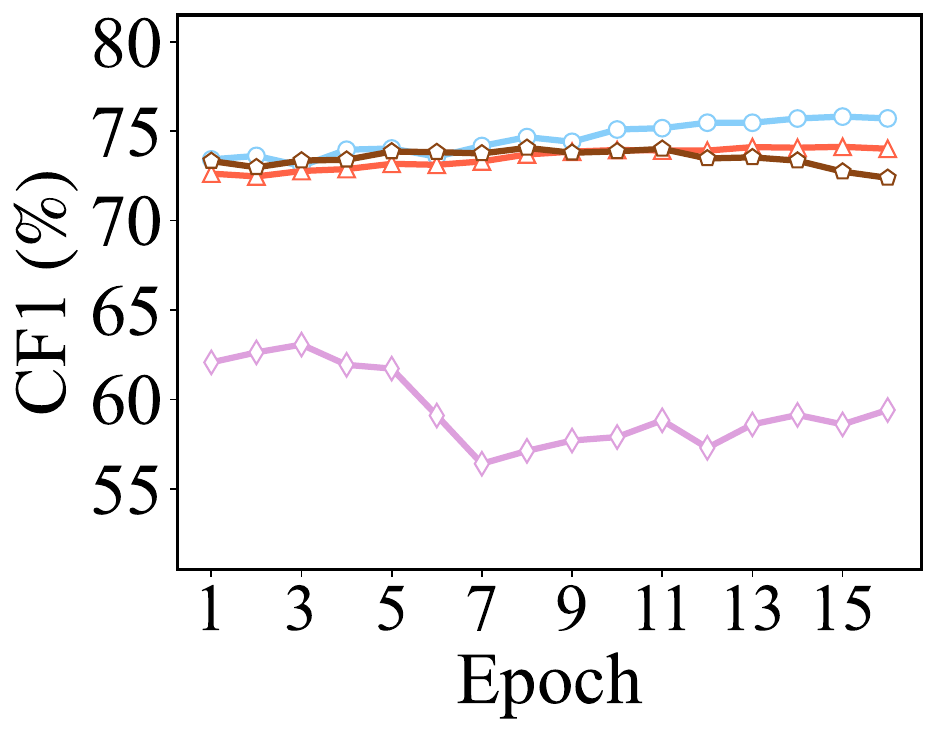}
		\caption{$p=0.1$.}
	\end{subfigure}
	\hfill
	\begin{subfigure}{0.24\linewidth}
		\includegraphics[width=0.99\linewidth]{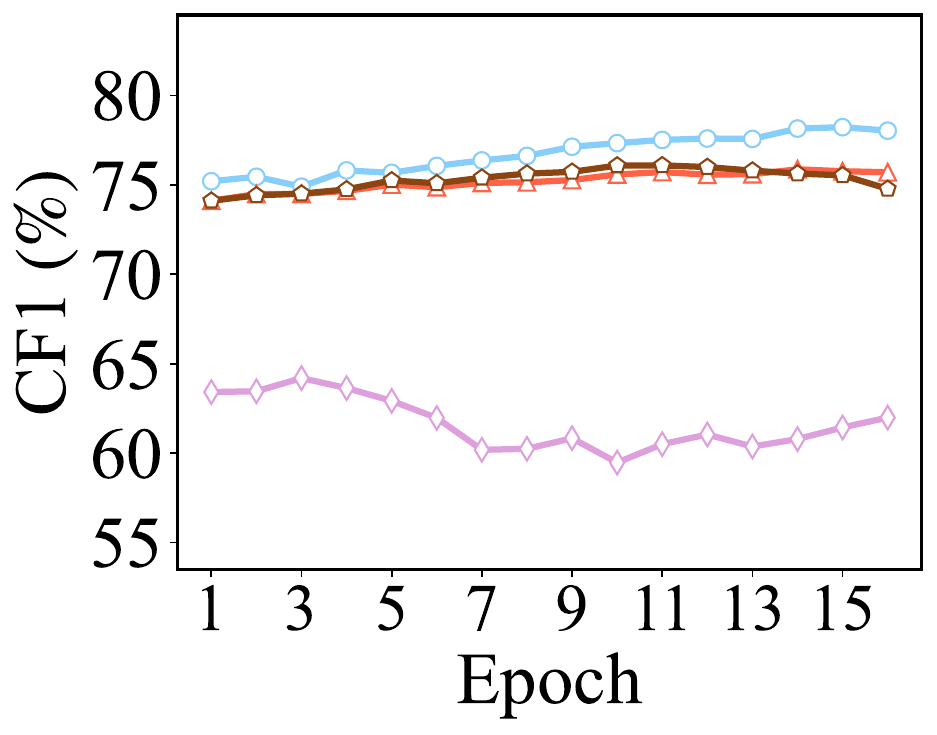}
		\caption{$p=0.15$.}
	\end{subfigure}
	\hfill
	\begin{subfigure}{0.24\linewidth}
		\includegraphics[width=0.99\linewidth]{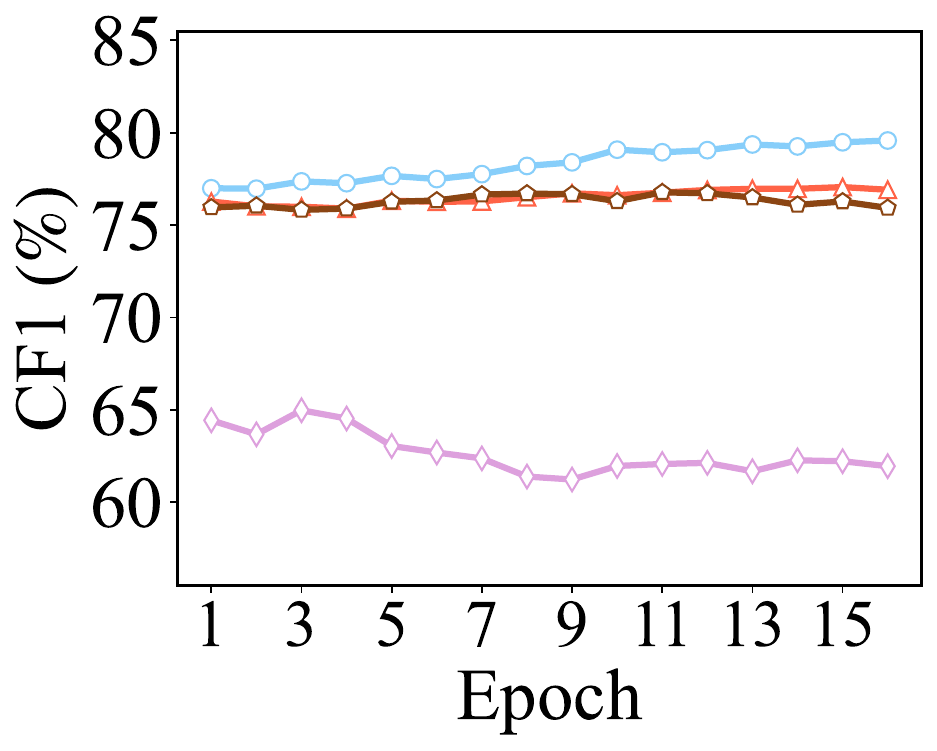}
		\caption{$p=0.2$.}
	\end{subfigure}
	
	\hfill
	\begin{subfigure}{0.24\linewidth}
		\includegraphics[width=0.99\linewidth]{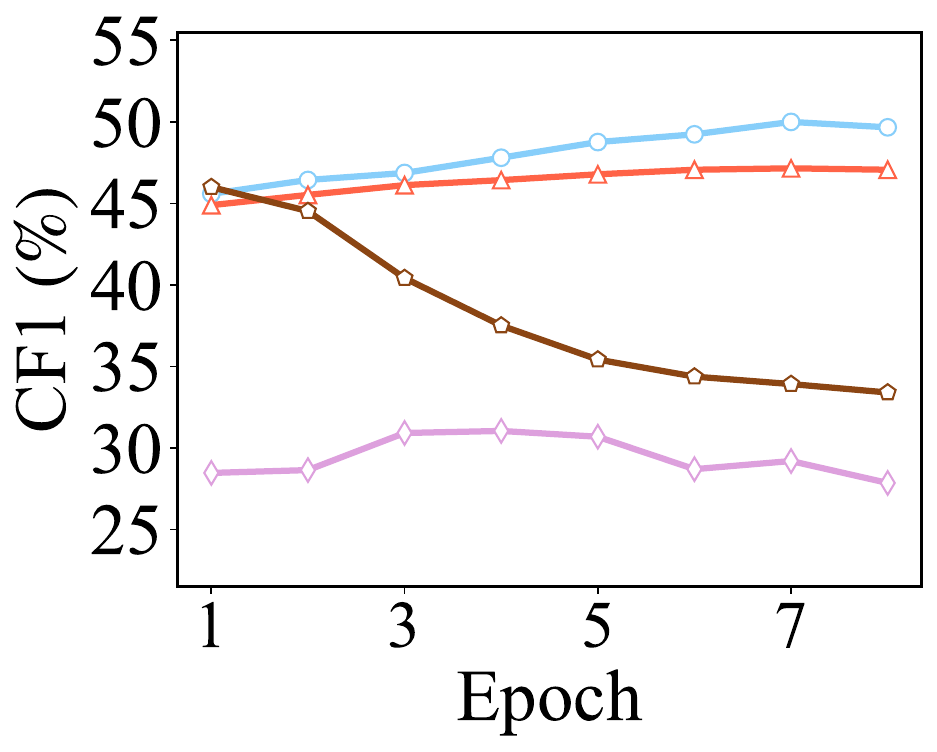}
		\caption{$p=0.05$.}
	\end{subfigure}
	\hfill
	\begin{subfigure}{0.24\linewidth}
		\includegraphics[width=0.99\linewidth]{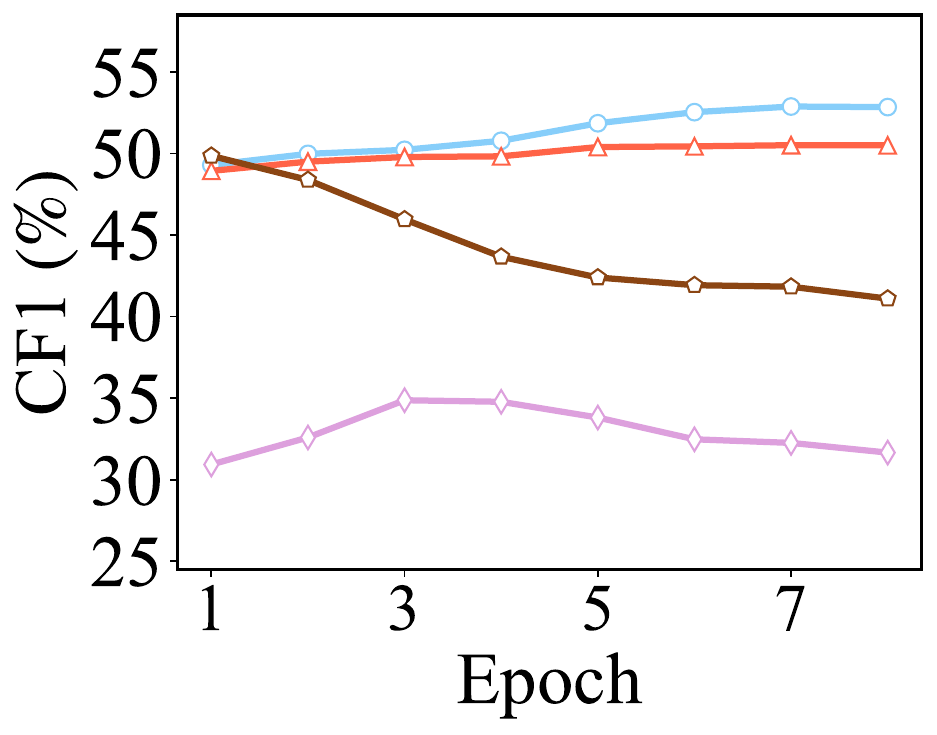}
		\caption{$p=0.1$.}
	\end{subfigure}
	\hfill
	\begin{subfigure}{0.24\linewidth}
		\includegraphics[width=0.99\linewidth]{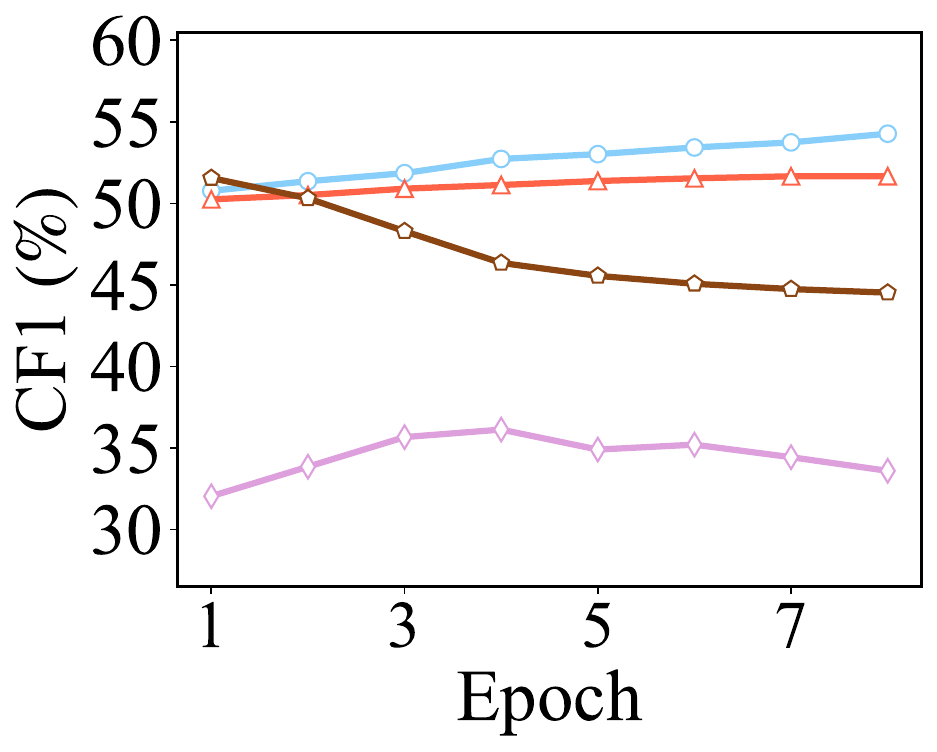}
		\caption{$p=0.15$.}
	\end{subfigure}
	\hfill
	\begin{subfigure}{0.24\linewidth}
		\includegraphics[width=0.99\linewidth]{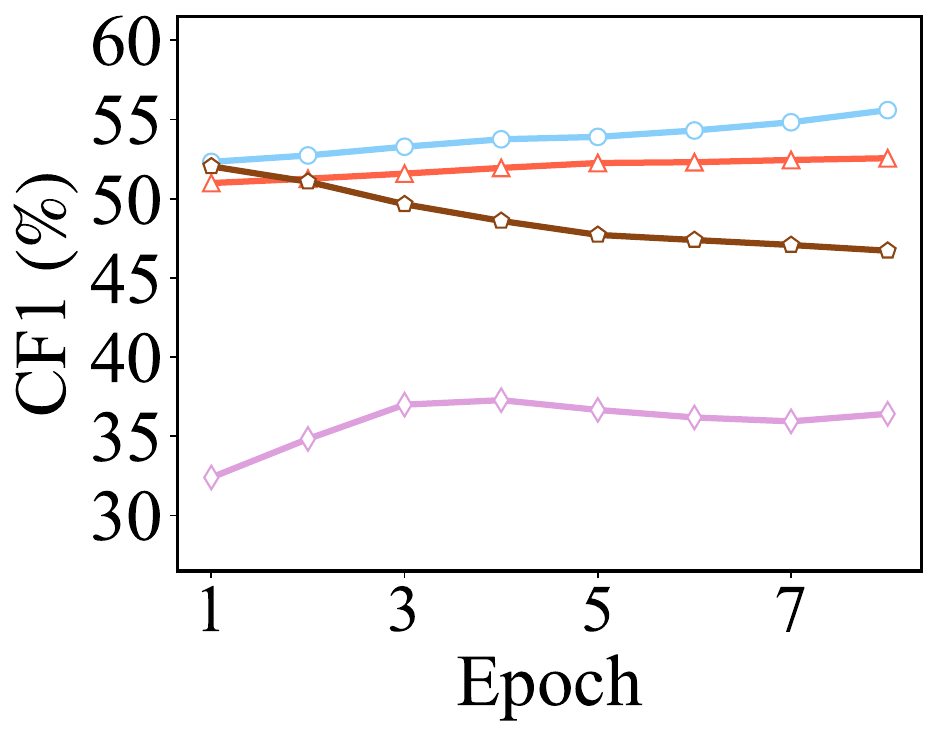}
		\caption{$p=0.2$.}
	\end{subfigure}
	\caption{The performance of pseudo-labeling on VOC (a-d) and NUS (e-h). }
	\label{fig:pseudo_voc_nus}
\end{figure}

\section{More Results of Pseudo-Labeling Performance}
\label{app:pseudo_label_performance}
In \Cref{fig:pseudo_voc_nus}, we also perform experiments to examine the quality of pseudo-labels generated by different methods on VOC and NUS in terms of CF1 score. We can observe phenomena similar to those described in the main text on COCO. For the relatively simpler VOC dataset among the three datasets, the first three methods perform comparably across most proportions, and our method outperforms the other two. Even on the challenging NUS dataset, our method exhibits commendable performance compared to the SOTA.

\section{Reproducibility and Resource Consumption}
\label{app:reproducibility}
To verify the reproducibility of our method, we conduct five runs using different random seeds ($\text{seed}=\{1,2,3,4,5\}$) and record the mean and standard deviation of our method's performance. In addition, we compare the results of the second-best method, CAP \cite{xie2023class}, after the same five runs. \Cref{tab:mean_std} presents a comparison between our method and CAP in terms of the mAP metric (with mean and standard deviation) across three datasets, showing that our method is not only reproducible but also superior to CAP. Furthermore, we compare the resource consumption of our method and CAP (also depicted in \Cref{tab:mean_std}). Although our method slightly exceeds CAP in terms of memory and time usage, this additional resource investment is acceptable given the significant performance improvement.

\begin{table}
    \caption{Mean and standard deviation of mAP(\%) in CAP and our method, on three datasets, along with the time/memory comparison. `Time' is the training time per epoch, including the process of threshold updating, `GPU' is the max memory allocated during training phase.}
    \label{tab:mean_std}
    \centering
    \begin{tabularx}{\textwidth}{l@{\hspace{5pt}}|@{\hspace{5pt}}XXX|@{\hspace{5pt}}XXX}
        \toprule
        Methods & \multicolumn{3}{c|@{\hspace{5pt}}}{CAP} & \multicolumn{3}{c}{Ours} \\
        \midrule
        Datasets & \multicolumn{1}{c}{VOC} & \multicolumn{1}{c}{COCO} & \multicolumn{1}{c|@{\hspace{5pt}}}{NUS} & \multicolumn{1}{c}{VOC} & \multicolumn{1}{c}{COCO} & \multicolumn{1}{c}{NUS} \\
        \midrule
        $p = 0.05$ & 77.15{\scriptsize ±0.58}             & 63.11{\scriptsize ±0.35}             & 45.30{\scriptsize ±0.30}           & 81.45{\scriptsize ±1.50}            & 70.15{\scriptsize ±0.48}             & 47.42{\scriptsize ±1.00}    \\
        $p = 0.10$ & 82.54{\scriptsize ±0.20}             & 67.96{\scriptsize ±0.32}             & 48.89{\scriptsize ±0.37}           & 85.65{\scriptsize ±0.92}            & 73.65{\scriptsize ±0.34}             & 51.01{\scriptsize ±0.43}     \\
        $p = 0.15$ & 83.95{\scriptsize ±0.24}            & 69.92{\scriptsize ±0.41}             & 50.53{\scriptsize ±0.54}            & 87.02{\scriptsize ±0.67}            & 75.18{\scriptsize ±0.31}             & 52.15{\scriptsize ±0.46}   \\
        $p = 0.20$ & 85.04{\scriptsize ±0.32}            & 71.23{\scriptsize ±0.42}             & 51.82{\scriptsize ±0.43}            & 87.83{\scriptsize ±0.38}            & 76.21{\scriptsize ±0.30}             & 53.37{\scriptsize ±0.43}    \\
        \midrule
        Time     & \multicolumn{1}{c}{0.9min}                  & \multicolumn{1}{c}{10.3min}               & \multicolumn{1}{c|@{\hspace{5pt}}}{12.2min}                 & \multicolumn{1}{c}{1.7min}                  & \multicolumn{1}{c}{21.8min}             & \multicolumn{1}{c}{38.9min}    \\
        \midrule
        GPU      & \multicolumn{3}{c|@{\hspace{5pt}}}{11.1G}                                                    & \multicolumn{3}{c}{14.2G}       \\
        \bottomrule
    \end{tabularx}
\end{table}


\section{More Ablation Studies}
\label{app:ablation}
\paragraph{The Study on D2L and MAT.}
In \Cref{tab:results_ablation_nus}, we report the results of ablation experiments on NUS, where the effectiveness of each component in our method is separately validated. Based on the baseline, we gradually introduce these components: metric-adaptive thresholding (MAT, in \Cref{sec:MAT}), correlative/discriminative features decoupling (CDD, in \Cref{sec:D2L}) and generation/utilization of pseudo-labels decoupling (GUD, in \Cref{sec:D2L}). At four different labeled proportions, each component exhibits positive effects.

\begin{table}[t]
	\caption{Mean average precision (mAP \%) of the baseline incorporated with different components, on the dataset NUS. The baseline here indicates the method CAP.}
	\label{tab:results_ablation_nus}
	\centering
	\begin{tabular}{c c c @{\hspace{5pt}}|@{\hspace{5pt}} c c c c}
		\toprule
		\multirow{2}*{MAT} & \multicolumn{2}{c}{D2L} & \multicolumn{4}{c}{NUS}  \\
		\cmidrule(lr){2-3}\cmidrule(lr){4-7} & CDD & GUD   & $p$=$0.05$ & $p$=$0.10$ & $p$=$0.15$ & $p$=$0.20$ \\ \midrule
		&		&		&	44.82	&	48.24	&	49.90	&	51.06	\\	
		\midrule
		\checkmark	&		&		&	45.26	&	48.88	&	50.23	&	51.20	\\	
		\checkmark	&	\checkmark	&		&	45.74	&	49.30	&	50.83	&	52.04	\\	
		\checkmark	&	\checkmark	&	\checkmark	&	46.86	&	50.25	&	51.61	&	52.64	\\	
		\bottomrule
	\end{tabular}
\end{table}

\begin{table}[t]
	\caption{Average per-class F1 (CF1 \%) score of each compared method. Bold represents the highest CF1. LL-* and Top-* select the best-performing method from their respective categories. The detailed method descriptions can be found in \ref{para:methods}.}
	\label{tab:cf1s}
	\centering
	\begin{tabularx}{\textwidth}{l@{\hspace{5pt}}|@{\hspace{5pt}}XXXXXXXXXXX}
		\multicolumn{12}{c}{CF1 score (\%) on VOC.} \\ 
		\toprule
		\multicolumn{1}{l}{Method} & \hspace{1pt}BCE   & \hspace{1pt}ASL   & LL-*  & \hspace{1pt}PLC   & Top-* & \hspace{2pt}IAT   & \hspace{-3pt}ADSH  & \hspace{2.5pt}FM    & \hspace{-4pt}DRML  & \hspace{1pt}CAP   & \hspace{1pt}Ours  \\
		\midrule
		$p=0.01$ & 19.29 & 36.42 & 37.78 & 40.62 & 36.93 & 38.91 & \textbf{45.60} & 44.20 & 36.15 & 44.54 & 41.15 \\
		$p=0.05$ & 54.00 & 59.76 & 62.00 & 62.20 & 63.33 & 60.18 & 60.80 & 61.18 & 52.99 & \textbf{69.86} & 68.50 \\
		$p=0.10$ & 62.86 & 62.70 & 65.75 & 66.81 & 67.24 & 65.18 & 64.58 & 65.93 & 62.41 & 75.63 & \textbf{75.94} \\
		$p=0.15$ & 64.14 & 66.17 & 67.40 & 67.37 & 67.68 & 66.04 & 66.38 & 66.72 & 63.10 & 77.09 & \textbf{77.14} \\
		$p=0.20$ & 63.96 & 64.47 & 67.45 & 66.80 & 67.71 & 66.97 & 66.34 & 67.56 & 63.35 & 77.88 & \textbf{79.37} \\
		\bottomrule
		\multicolumn{12}{c}{\ } \\ 
	\end{tabularx}
	\begin{tabularx}{\textwidth}{l@{\hspace{5pt}}|@{\hspace{5pt}}XXXXXXXXXXX}
		\multicolumn{12}{c}{CF1 score (\%) on COCO.}\\
		\toprule
		\multicolumn{1}{l}{Method} & \hspace{1pt}BCE   & \hspace{1pt}ASL   & LL-*  & \hspace{1pt}PLC   & Top-* & \hspace{2pt}IAT   & \hspace{-3pt}ADSH  & \hspace{2.5pt}FM    & \hspace{-4pt}DRML  & \hspace{1pt}CAP   & \hspace{1pt}Ours  \\
		\midrule
		$p=0.01$ & 41.19 & 41.08 & 42.80 & 44.36 & 47.19 & 42.03 & 44.42 & 43.28 & 33.53 & 52.70 & \textbf{55.32} \\
		$p=0.05$ & 51.52 & 51.05 & 53.38 & 53.20 & 51.69 & 51.67 & 53.20 & 51.49 & 46.81 & 60.66 & \textbf{65.65} \\
		$p=0.10$ & 54.11 & 53.46 & 56.59 & 55.92 & 54.97 & 55.40 & 56.17 & 54.17 & 48.71 & 64.11 & \textbf{68.70} \\
		$p=0.15$ & 55.48 & 55.01 & 57.75 & 57.99 & 56.60 & 56.55 & 57.65 & 55.66 & 49.89 & 65.40 & \textbf{69.83} \\
		$p=0.20$ & 56.44 & 55.78 & 58.72 & 59.07 & 57.63 & 57.51 & 58.40 & 56.72 & 51.08 & 66.30 & \textbf{70.74} \\
		\bottomrule
		\multicolumn{12}{c}{\ } \\ 
	\end{tabularx}
	\begin{tabularx}{\textwidth}{l@{\hspace{5pt}}|@{\hspace{5pt}}XXXXXXXXXXX}
		\multicolumn{12}{c}{CF1 score (\%) on NUS.} \\
		\toprule
		\multicolumn{1}{l}{Method} & \hspace{1pt}BCE   & \hspace{1pt}ASL   & LL-*  & \hspace{1pt}PLC   & Top-* & \hspace{2pt}IAT   & \hspace{-3pt}ADSH  & \hspace{2.5pt}FM    & \hspace{-4pt}DRML  & \hspace{1pt}CAP   & \hspace{1pt}Ours  \\
		\midrule
		$p=0.01$ & 23.68 & 23.22 & 20.30 & 23.42 & 22.39 & 22.78 & 30.19 & 26.80 & 16.36 & 28.81 & \textbf{42.17} \\
		$p=0.05$ & 33.57 & 32.83 & 31.70 & 31.75 & 34.57 & 32.10 & 36.29 & 33.24 & 25.48 & 47.14 & \textbf{47.47} \\
		$p=0.10$ & 36.75 & 34.35 & 35.17 & 34.96 & 37.24 & 34.84 & 38.20 & 36.15 & 28.05 & 49.94 & \textbf{50.77} \\
		$p=0.15$ & 38.33 & 35.38 & 35.81 & 36.61 & 38.55 & 35.94 & 37.79 & 37.34 & 28.95 & 51.14 & \textbf{51.35} \\
		$p=0.20$ & 39.59 & 36.47 & 37.21 & 38.27 & 39.70 & 37.16 & 38.68 & 38.65 & 30.31 & \textbf{52.37} & 52.31 \\
		\bottomrule
	\end{tabularx}
\end{table}

\begin{table}[t]
	\caption{Overall F1 (OF1 \%) score of each compared method. Bold represents the highest OF1. LL-* and Top-* select the best-performing method from their respective categories. The detailed method descriptions can be found in \ref{para:methods}.}
	\label{tab:of1s}
	\centering
	\begin{tabularx}{\textwidth}{l@{\hspace{5pt}}|@{\hspace{5pt}}XXXXXXXXXXX}
		\multicolumn{12}{c}{OF1 score (\%) on VOC.} \\ 
		\toprule
		\multicolumn{1}{l}{Method} & \hspace{1pt}BCE   & \hspace{1pt}ASL   & LL-*  & \hspace{1pt}PLC   & Top-* & \hspace{2pt}IAT   & \hspace{-3pt}ADSH  & \hspace{2.5pt}FM    & \hspace{-4pt}DRML  & \hspace{1pt}CAP   & \hspace{1pt}Ours  \\
		\midrule
		$p=0.01$ & 31.55 & 43.63 & 42.33 & 41.93 & 43.21 & 45.48 & \textbf{52.79} & 49.93 & 46.12 & 33.57 & 35.63 \\
		$p=0.05$ & 60.63 & 63.47 & 65.17 & 64.46 & 65.16 & 63.95 & 64.69 & 64.99 & 56.40 & \textbf{73.98} & 72.23 \\
		$p=0.10$ & 65.36 & 66.11 & 68.11 & 67.73 & 68.00 & 67.63 & 67.54 & 67.92 & 61.01 & 78.39 & \textbf{79.38} \\
		$p=0.15$ & 66.34 & 66.83 & 69.07 & 68.95 & 68.94 & 68.55 & 68.84 & 68.77 & 62.25 & 79.83 & \textbf{80.54} \\
		$p=0.20$ & 66.95 & 67.25 & 69.37 & 69.04 & 69.22 & 69.12 & 69.13 & 69.42 & 63.55 & 80.80 & \textbf{82.44} \\
		\bottomrule
		\multicolumn{12}{c}{\ } \\ 
	\end{tabularx}
	\begin{tabularx}{\textwidth}{l@{\hspace{5pt}}|@{\hspace{5pt}}XXXXXXXXXXX}
		\multicolumn{12}{c}{OF1 score (\%) on COCO.}\\
		\toprule
		\multicolumn{1}{l}{Method} & \hspace{1pt}BCE   & \hspace{1pt}ASL   & LL-*  & \hspace{1pt}PLC   & Top-* & \hspace{2pt}IAT   & \hspace{-3pt}ADSH  & \hspace{2.5pt}FM    & \hspace{-4pt}DRML  & \hspace{1pt}CAP   & \hspace{1pt}Ours  \\
		\midrule
		$p=0.01$ & 49.72 & 50.28 & 51.20 & 51.91 & 53.64 & 51.55 & 51.76 & 52.00 & 45.76 & 59.99 & \textbf{62.67} \\
		$p=0.05$ & 57.47 & 57.37 & 58.96 & 58.46 & 57.68 & 58.37 & 58.16 & 57.88 & 52.88 & 66.09 & \textbf{70.88} \\
		$p=0.10$ & 59.68 & 59.67 & 60.87 & 60.82 & 60.12 & 60.90 & 60.47 & 60.14 & 54.59 & 68.85 & \textbf{73.50} \\
		$p=0.15$ & 60.76 & 60.98 & 61.84 & 62.22 & 61.33 & 62.12 & 61.64 & 61.33 & 55.66 & 69.94 & \textbf{74.33} \\
		$p=0.20$ & 61.48 & 61.58 & 62.51 & 63.01 & 62.17 & 62.73 & 62.32 & 62.01 & 55.89 & 70.71 & \textbf{75.10} \\
		\bottomrule
		\multicolumn{12}{c}{\ } \\ 
	\end{tabularx}
	\begin{tabularx}{\textwidth}{l@{\hspace{5pt}}|@{\hspace{5pt}}XXXXXXXXXXX}
		\multicolumn{12}{c}{OF1 score (\%) on NUS.} \\
		\toprule
		\multicolumn{1}{l}{Method} & \hspace{1pt}BCE   & \hspace{1pt}ASL   & LL-*  & \hspace{1pt}PLC   & Top-* & \hspace{2pt}IAT   & \hspace{-3pt}ADSH  & \hspace{2.5pt}FM    & \hspace{-4pt}DRML  & \hspace{1pt}CAP   & \hspace{1pt}Ours  \\
		\midrule
		$p=0.01$ & 47.30 & 46.89 & 36.14 & 48.24 & 39.03 & 44.95 & 47.84 & \textbf{48.47} & 42.38 & 35.26 & 40.65 \\
		$p=0.05$ & 50.05 & 50.45 & 50.50 & 51.28 & 50.77 & 50.46 & 50.94 & 50.72 & 46.93 & \textbf{66.92} & 66.23 \\
		$p=0.10$ & 50.99 & 51.36 & 51.48 & 52.17 & 51.75 & 51.45 & 51.86 & 51.68 & 48.07 & 68.09 & \textbf{68.50} \\
		$p=0.15$ & 51.58 & 51.95 & 51.99 & 52.59 & 52.17 & 52.01 & 52.37 & 52.06 & 48.72 & 68.62 & \textbf{68.74} \\
		$p=0.20$ & 51.72 & 52.22 & 52.30 & 52.91 & 52.37 & 52.36 & 52.67 & 52.40 & 49.06 & \textbf{69.23} & 69.15 \\
		\bottomrule
	\end{tabularx}
\end{table}

\begin{figure}[h]
	\centering
	\begin{subfigure}{0.55\linewidth}
			\includegraphics[width=1.0\linewidth]{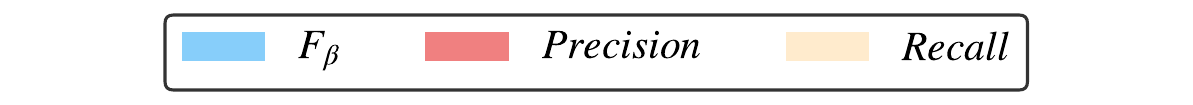}
	\end{subfigure}

	\hfill
	\begin{subfigure}{0.32\linewidth}\hspace{2pt}
		\includegraphics[width=0.8\linewidth]{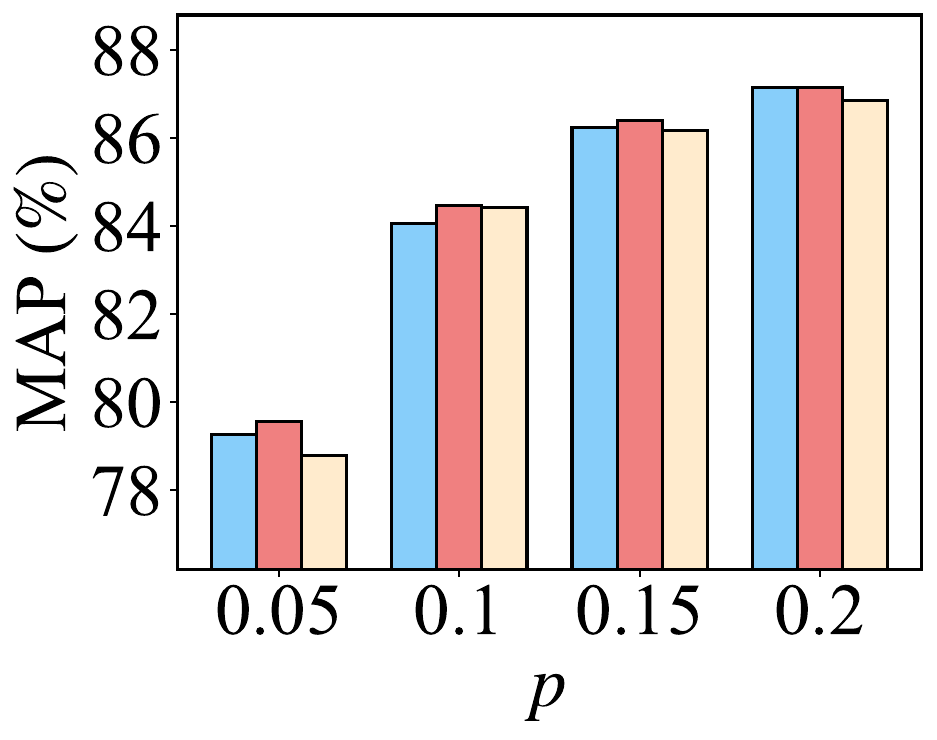}
		\caption{VOC.}
		\label{fig:voc_mat_supp}
	\end{subfigure}
	\hfill
	\begin{subfigure}{0.32\linewidth}\hspace{2pt}
		\includegraphics[width=0.8\linewidth]{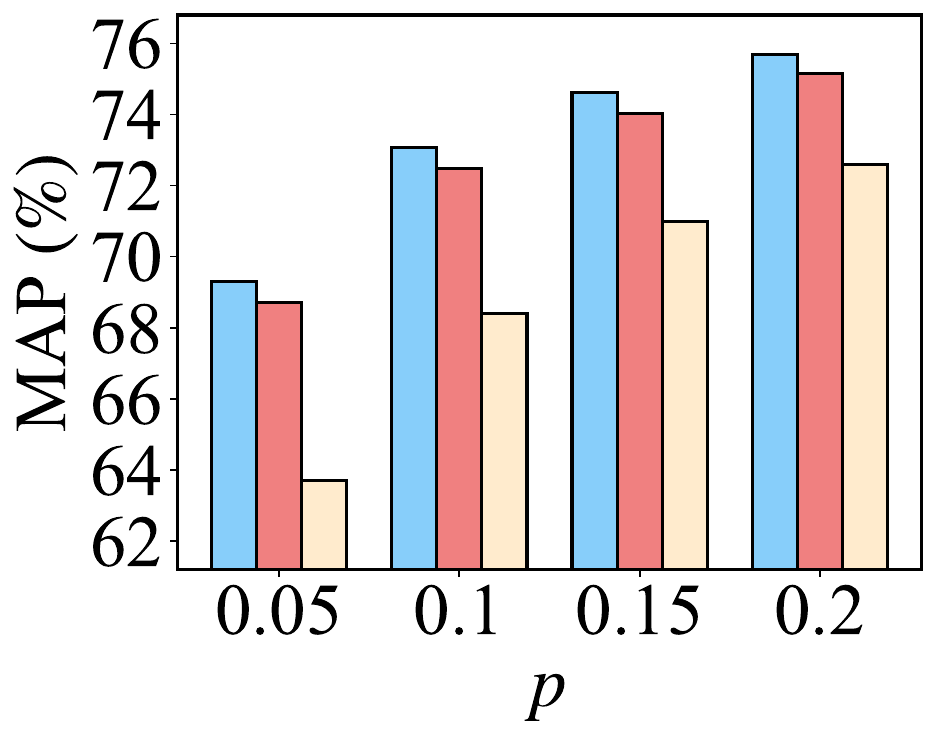}
		\caption{COCO.}
		\label{fig:coco_mat_supp}
	\end{subfigure}
	\hfill
	\begin{subfigure}{0.32\linewidth}\hspace{2pt}
		\includegraphics[width=0.8\linewidth]{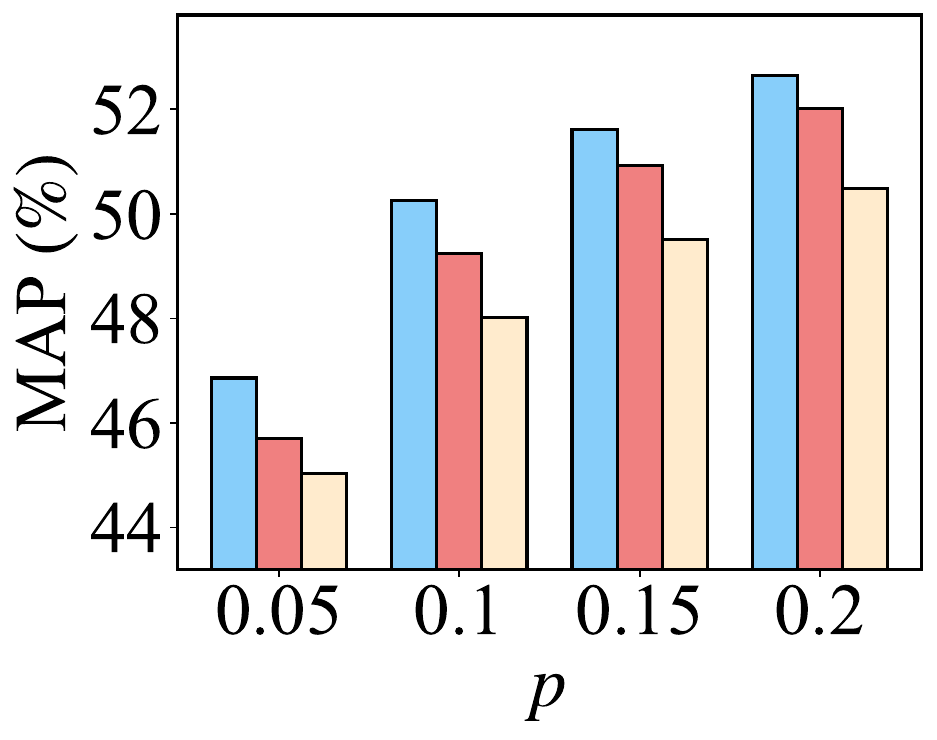}
		\caption{NUS.}
		\label{fig:nus_mat_supp}
	\end{subfigure}
	
	\caption{The analyses of metric function $\mathcal{M}(\cdot,\cdot)$ in MAT on three datasets. }
	\label{fig:metric}
\end{figure}

\begin{figure}[h]
	\centering
	\begin{subfigure}{0.55\linewidth}
		\includegraphics[width=1.0\linewidth]{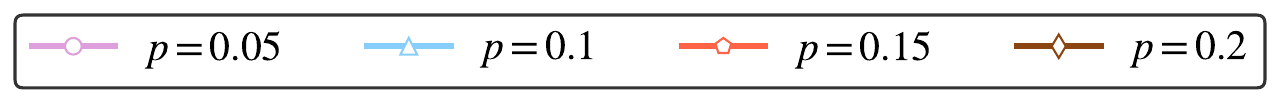}
	\end{subfigure}

	\hfill
	\begin{subfigure}{0.32\linewidth}\hspace{2pt}
		\includegraphics[width=0.8\linewidth]{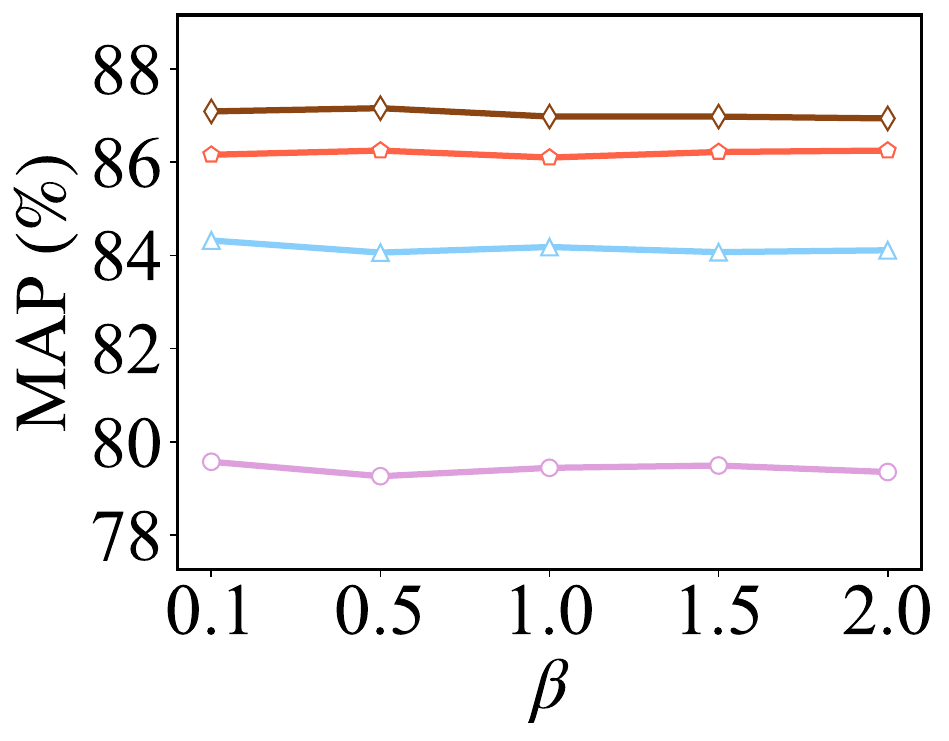}
		\caption{VOC.}
		\label{fig:voc_beta_supp}
	\end{subfigure}
	\hfill
	\begin{subfigure}{0.32\linewidth}\hspace{2pt}
		\includegraphics[width=0.8\linewidth]{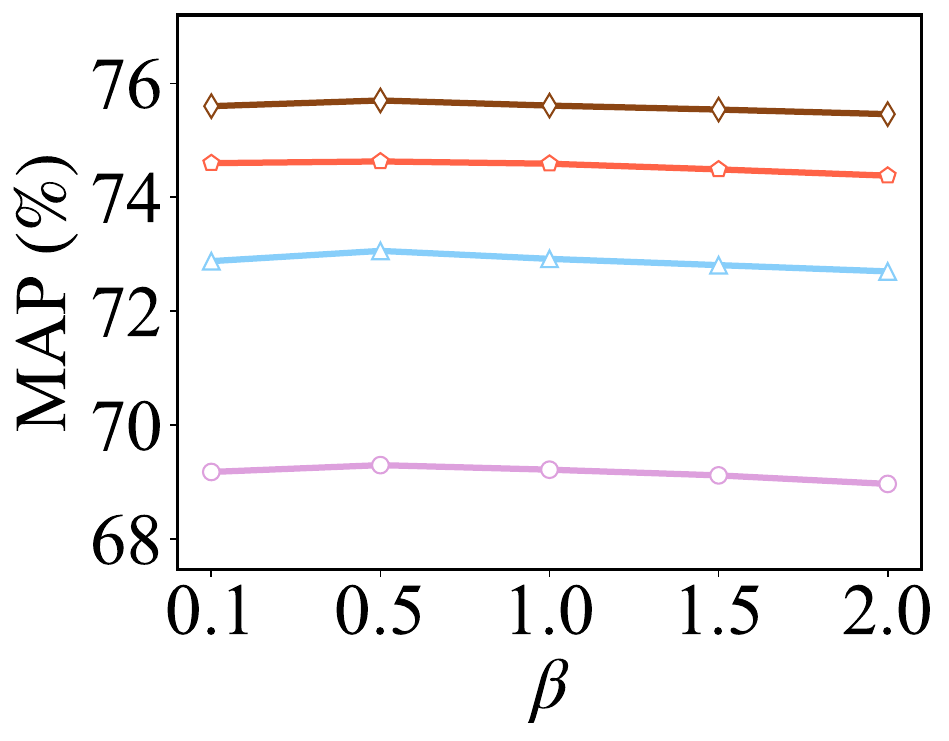}
		\caption{COCO.}
		\label{fig:coco_beta_supp}
	\end{subfigure}
	\hfill
	\begin{subfigure}{0.32\linewidth}\hspace{2pt}
		\includegraphics[width=0.8\linewidth]{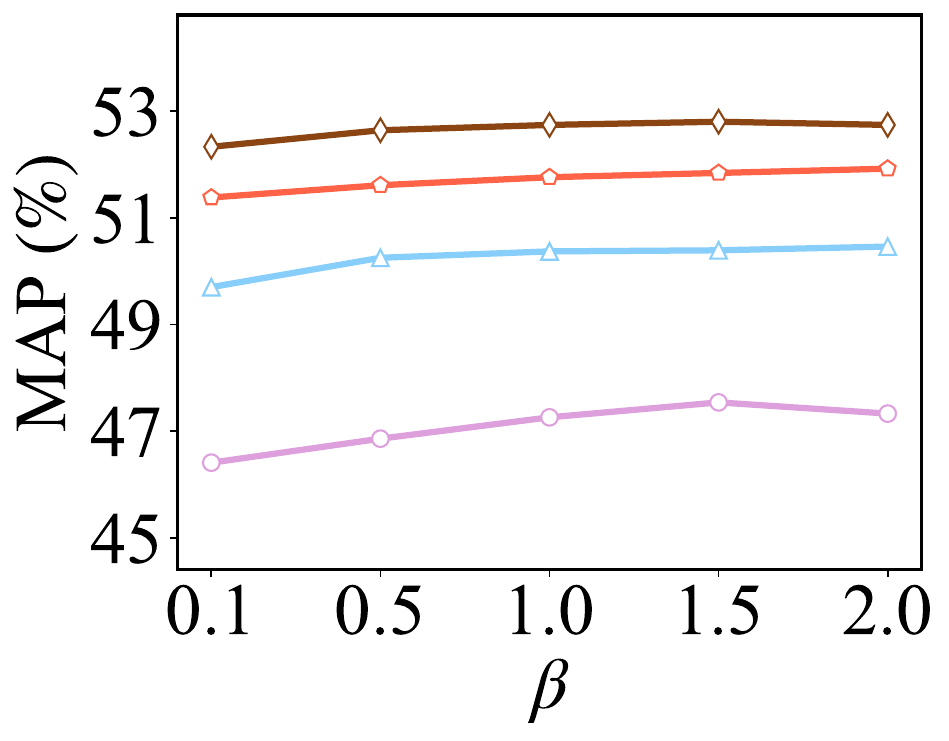}
		\caption{NUS.}
		\label{fig:nus_beta_supp}
	\end{subfigure}
	
	\caption{The analyses of value $\beta$ in metric function $F_\beta$ on three datasets. }
	\label{fig:beta}
\end{figure}

\paragraph{The Study on Metric Function.}
\Cref{fig:metric,fig:beta} present the analyses of parameters, including the metric function $\mathcal{M}(\cdot,\cdot)$ and the value $\beta$ in metric $F_{\beta}$, across three datasets. For VOC, the three metric functions perform comparably across the four labeled proportions and the insensitivity of $\beta$ in $F_{\beta}$ remains consistent with COCO. For NUS, choosing $F_\beta$ appears to be a more suitable metric. However, in cases of low labeled proportions, a higher value of $\beta$ needs to be selected as performance tends to increase with the increase in $\beta$.

\begin{figure}[h]
	\centering
	\begin{subfigure}{0.55\linewidth}
		\includegraphics[width=1.0\linewidth]{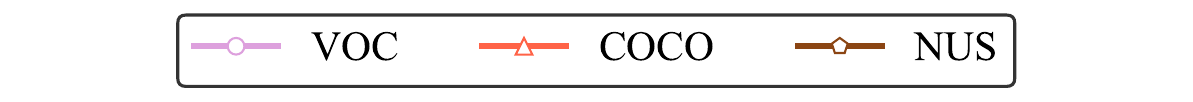}
	\end{subfigure}
	
	\hfill
	\begin{subfigure}{0.24\linewidth}
		\includegraphics[width=0.99\linewidth]{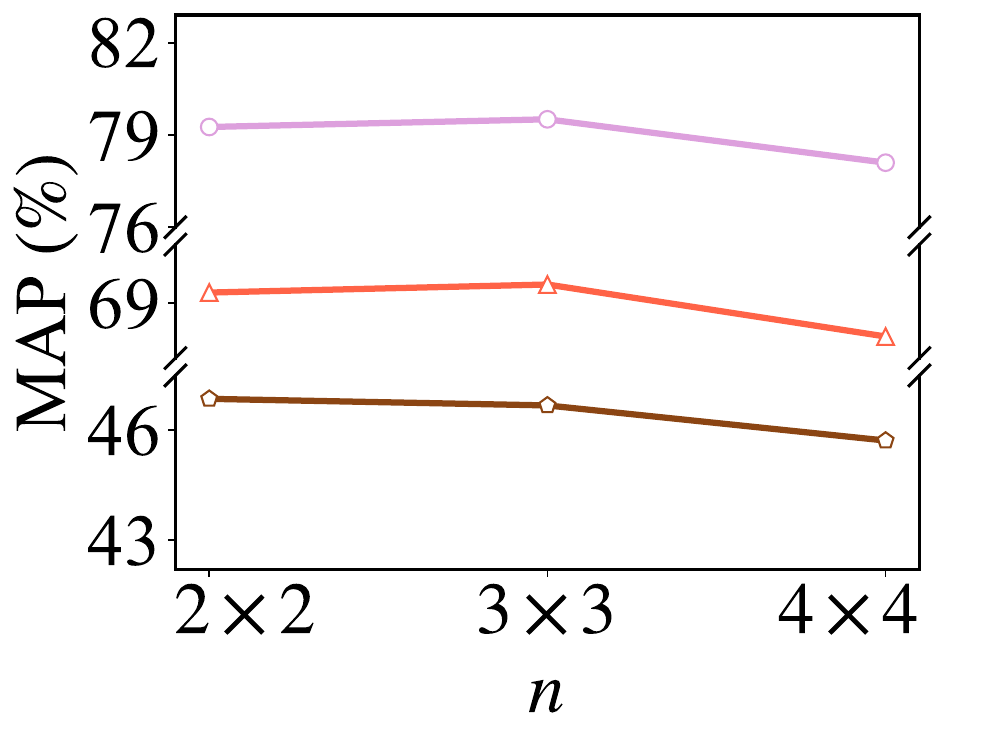}
		\caption{$p=0.05$.}
		
	\end{subfigure}
	\hfill
	\begin{subfigure}{0.24\linewidth}
		\includegraphics[width=0.99\linewidth]{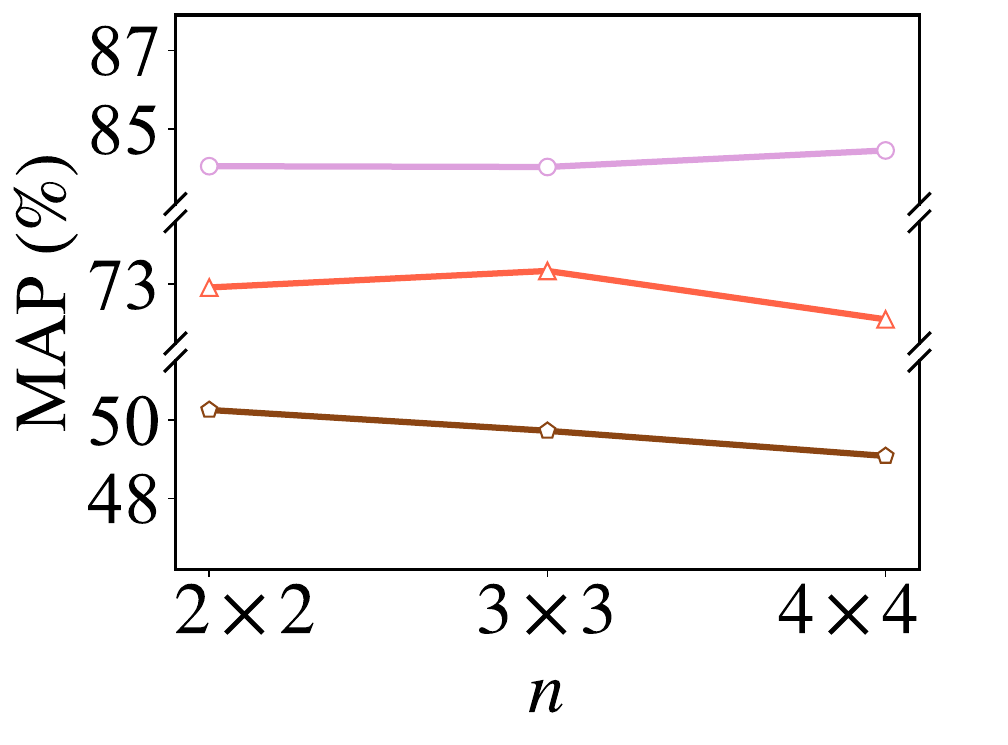}
		\caption{$p=0.1$.}
		
	\end{subfigure}
	\hfill
	\begin{subfigure}{0.24\linewidth}
		\includegraphics[width=0.99\linewidth]{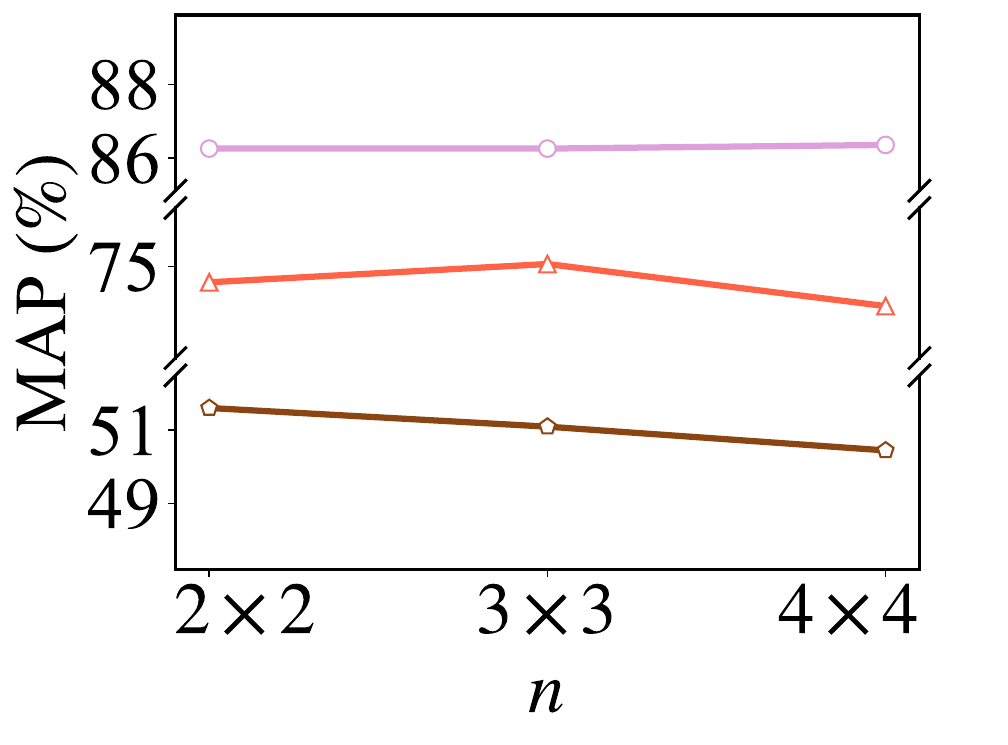}
		\caption{$p=0.15$.}
		
	\end{subfigure}
	\hfill
	\begin{subfigure}{0.24\linewidth}
		\includegraphics[width=0.99\linewidth]{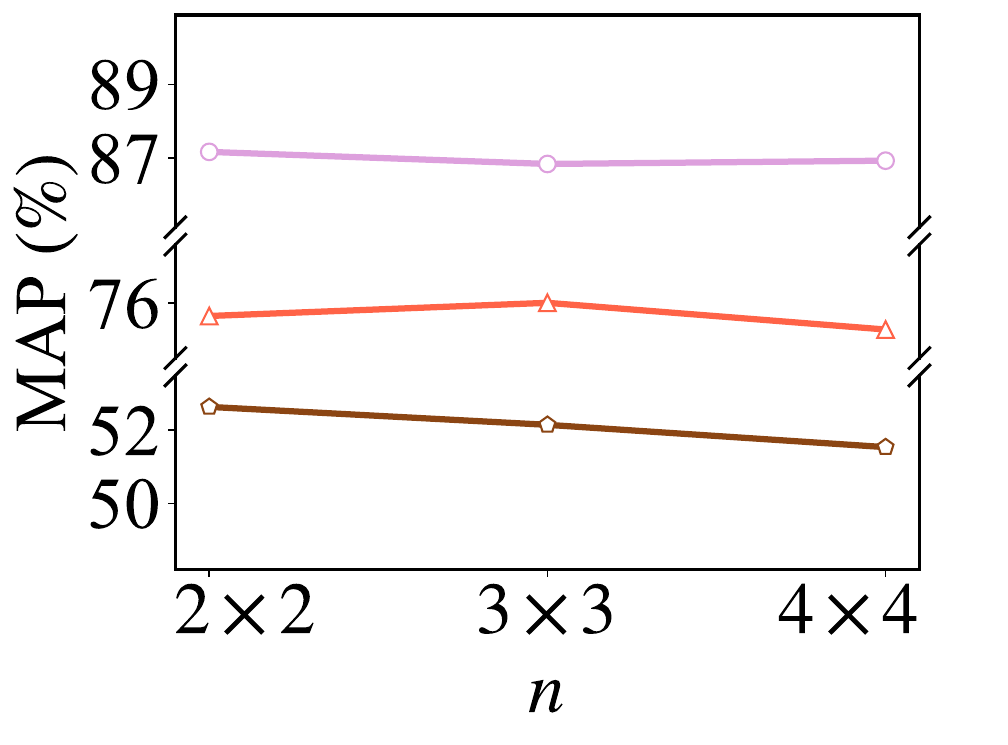}
		\caption{$p=0.2$.}
		
	\end{subfigure}

	\caption{The analyses of number of patches $n$ on three datasets. }
	\label{fig:0.05-0.2_n}
\end{figure}

\begin{figure}[h]
	\centering
	\begin{subfigure}{0.55\linewidth}
		\includegraphics[width=1.0\linewidth]{fig/legend_grid.pdf}
	\end{subfigure}
	
	\hfill
	\begin{subfigure}{0.23\linewidth}\hspace{-7pt}
		\includegraphics[width=0.99\linewidth]{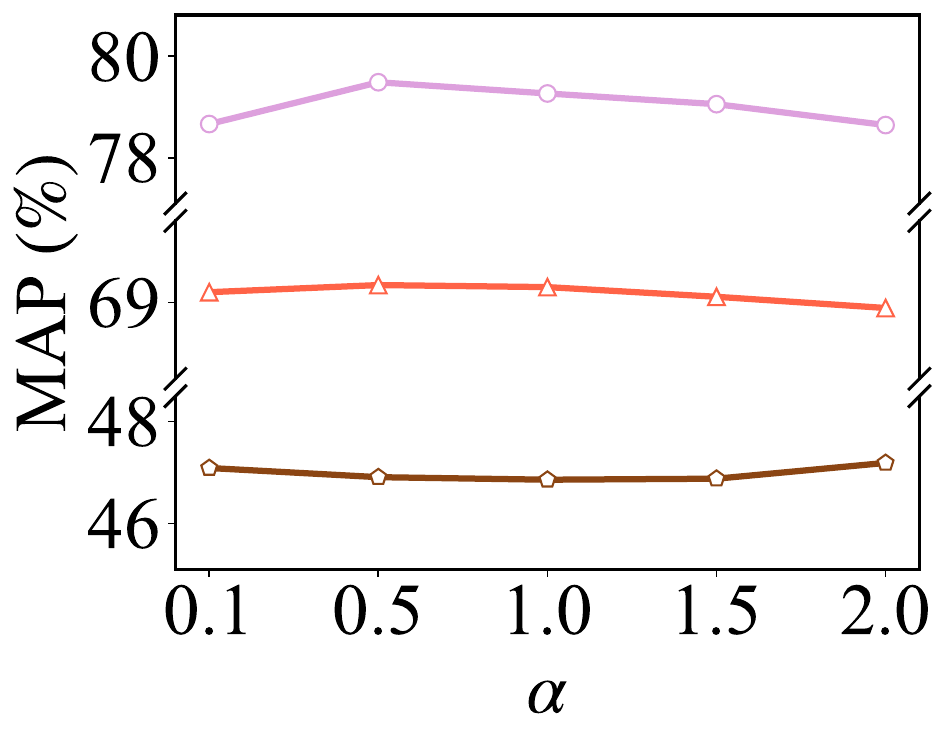}
		\caption{$p=0.05$.}
	\end{subfigure}
	\hfill
	\begin{subfigure}{0.23\linewidth}\hspace{-7pt}
		\includegraphics[width=0.99\linewidth]{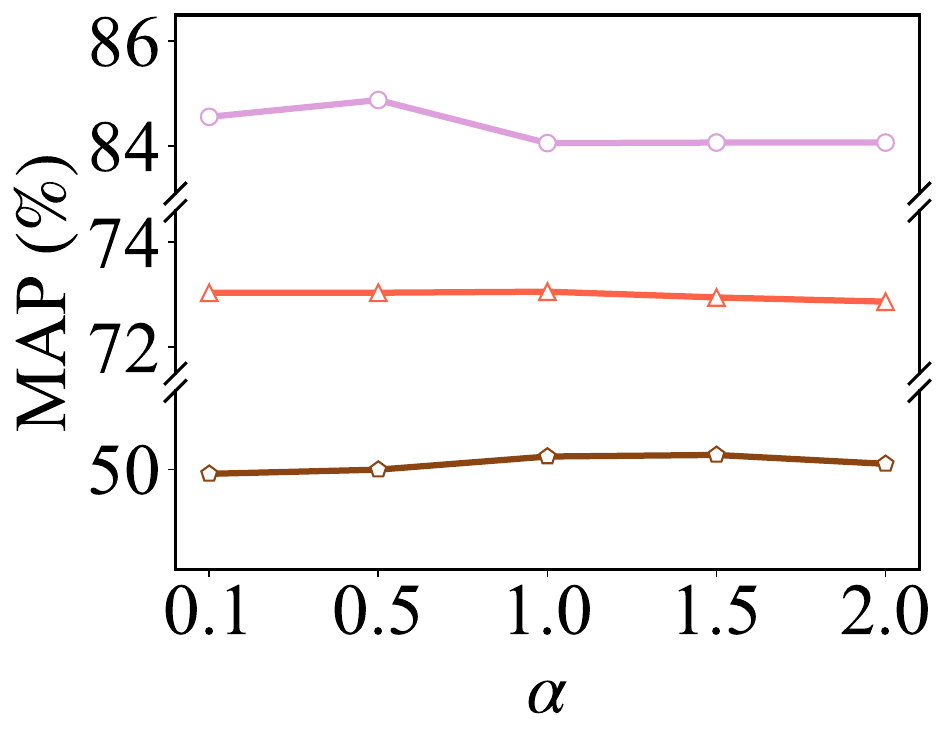}
		\caption{$p=0.1$.}
	\end{subfigure}
	\hfill
	\begin{subfigure}{0.23\linewidth}\hspace{-7pt}
		\includegraphics[width=0.99\linewidth]{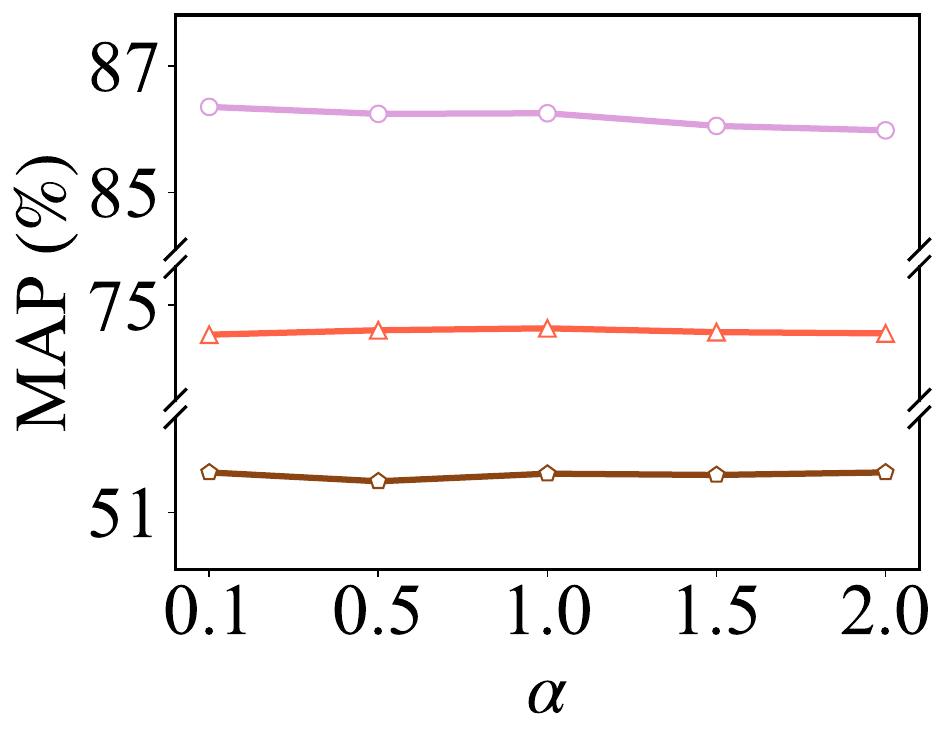}
		\caption{$p=0.15$.}
	\end{subfigure}
	\hfill
	\begin{subfigure}{0.23\linewidth}\hspace{-7pt}
		\includegraphics[width=0.99\linewidth]{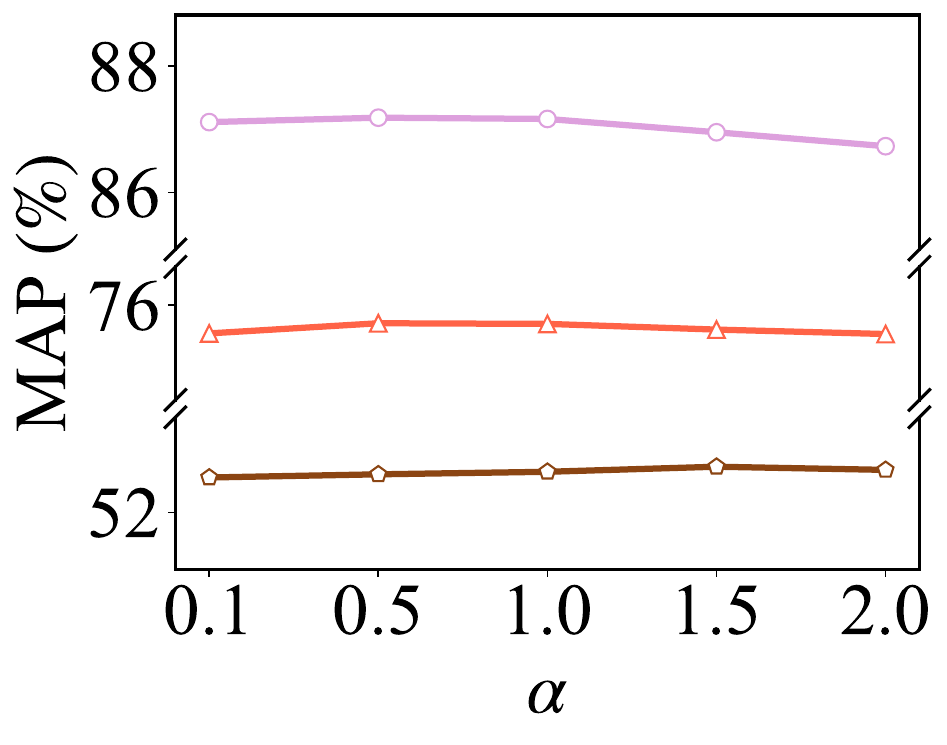}
		\caption{$p=0.2$.}
	\end{subfigure}
	
	\caption{The analyses of temperature $\alpha$ on three datasets. }
	\label{fig:0.05-0.2_alpha}
\end{figure}

\paragraph{Parameter Sensitivity Analyses.}
In \Cref{fig:0.05-0.2_n,fig:0.05-0.2_alpha}, we demonstrate the performance variation with the parameters $n$ and $\alpha$ within the range $\{2\times2, 3\times3, 4\times4\}$ and $\{0.1,0.5,1.0,1.5,2.0\}$, respectively. For the sake of presentation, we include figures from the main text where $p=0.05$ alongside figures with $p=\{0.1, 0.15, 0.2\}$ that were not previously displayed. For parameter $n$, considering all three datasets, we recommend using $n=2\times 2$ for cropping since it not only saves some computational costs but also achieves decent performance. For parameter $\alpha$, our method is generally insensitive to it. So, we use $\alpha=1.0$ in all experiments for simplicity.

\section{More Results of Additional Evaluation Metrics}
\label{app:cf1_of1}
In \Cref{tab:cf1s,tab:of1s}, we present additional comparative experimental results that were not reported in the main text. This includes the results of two newly introduced metrics, average per-class F1 score (CF1) and overall F1 score (OF1), across three datasets and five annotation ratios. Specifically, these two metrics can be computed as follow:
\begin{equation}\nonumber
	\label{eq:cf1_of1}
	\text{CF1}=\frac{2\times \text{CP} \times \text{CR}}{\text{CP} + \text{CR}}, \ \ \ \text{OF1}=\frac{2\times \text{OP} \times \text{OR}}{\text{OP} + \text{OR}},
\end{equation}
and
\begin{equation}\nonumber
	\begin{aligned}
		\label{eq:cp_op_cr_or}
		&\text{CP}=\frac{1}{K}\sum_{k}\frac{N_k^{TP}}{N_k^{TP}+N_k^{FP}}, &\text{OP}=\frac{\sum_{k}N_k^{TP}}{\sum_{k}(N_k^{TP}+N_k^{FP})},\\
		&\text{CR}=\frac{1}{K}\sum_{k}\frac{N_k^{TP}}{N_k^{TP}+N_k^{FN}}, &\text{OR}=\frac{\sum_{k}N_k^{TP}}{\sum_{k}(N_k^{TP}+N_k^{FN})},
	\end{aligned}
\end{equation}
where CP, CR are average per-class precision, recall, and OP, OR are overall precision, recall. According to the confusion matrix, $\{N_k^{TP}, N_k^{FP}, N_k^{TN}, N_k^{FN}\}$ indicate the number of true positive, false positive, true negative, false negative for the $k$-th class. The superiority of our approach is validated by these experimental results.

\end{document}